\documentclass[format=sigconf, nonacm=false, anonymous=false, screen=false, review=false]{acmart}

\AtBeginDocument{%
  \providecommand\BibTeX{{%
    \normalfont B\kern-0.5em{\scshape i\kern-0.25em b}\kern-0.8em\TeX}}}

\usepackage{graphicx}
\usepackage[spanish,ngerman,main=english]{babel}
\usepackage[utf8]{inputenc}
\usepackage{amsfonts}
\usepackage{amsmath}

\usepackage{algorithmic}
\usepackage[linesnumbered,ruled,vlined]{algorithm2e}

\SetCommentSty{mycommfont}
\SetKwInput{KwInput}{Input}                % Set the Input
\SetKwInput{KwOutput}{Output}              % set the Output
\SetKwInput{KwRequire}{Require}              % set the Require
\usepackage{array}
\usepackage{soul}
\usepackage{multirow}
\usepackage{xtab,booktabs,rotating}
\usepackage[caption=true,font=footnotesize]{subfig}
\usepackage{xhfill}
\usepackage{caption}
\usepackage[export]{adjustbox}
\usepackage{makecell}
\usepackage{xcolor}         % colors
\usepackage{url}
\usepackage{hyperref}
\usepackage[colorinlistoftodos]{todonotes}
\usepackage{microtype}        % Improved Tracking and Kerning

\usepackage{minitoc}
\usepackage{enumitem}

\definecolor{stelios_colour}{RGB}{200, 238, 200}

\newcolumntype{s}{>{\columncolor{Lavender}} c}
\newcolumntype{d}{>{\columncolor{Thistle3}} c}
\newcolumntype{f}{>{\columncolor{LightPink1}} c}

\newcommand{\sysname}{LifeLearner}

\newcommand{\repourl}{https://github.com/theyoungkwon/LifeLearner}

\makeatletter
  \if@ACM@anonymous
    \renewcommand{\repourl}{https://anonymous.4open.science/r/LifeLearner-B189}
  \fi
\makeatother

\copyrightyear{2023}
\acmYear{2023}
\setcopyright{rightsretained}
\acmConference[SenSys '23]{The 21st ACM Conference on Embedded Networked Sensor Systems}{November 12--17, 2023}{Istanbul, Turkiye}
\acmBooktitle{The 21st ACM Conference on Embedded Networked Sensor Systems (SenSys '23), November 12--17, 2023, Istanbul, Turkiye}\acmDOI{10.1145/3625687.3625804}
\acmISBN{979-8-4007-0414-7/23/11}

\begin{document}

\title{LifeLearner: Hardware-Aware Meta Continual Learning System for Embedded Computing Platforms}

%through Product Quantization

\author{Young D. Kwon}
\affiliation{%
  \institution{University of Cambridge}
  \country{United Kingdom}
}
\email{ydk21@cam.ac.uk}

\author{Jagmohan Chauhan}
\affiliation{%
  \institution{University of Southampton}
  \country{United Kingdom}
}
\email{J.Chauhan@soton.ac.uk}

\author{Hong Jia}
\affiliation{%
  \institution{University of Cambridge}
  \country{United Kingdom}
}
\email{hj359@cam.ac.uk}

\author{Stylianos I. Venieris}
\affiliation{%
  \institution{Samsung AI Center, Cambridge}
  \country{United Kingdom}
}
\email{s.venieris@samsung.com}

\author{Cecilia Mascolo}
\affiliation{%
  \institution{University of Cambridge}
  \country{United Kingdom}
}
\email{cm542@cam.ac.uk}

\renewcommand{\shortauthors}{Kwon et al.}

\begin{abstract}

Continual Learning (CL) allows applications such as user personalization and household robots to learn on the fly and adapt to context. This is an important feature when context, actions, and users change. However, enabling CL on resource-constrained embedded systems is challenging due to the limited labeled data, memory, and computing capacity.

In this paper, we propose \sysname, a hardware-aware meta continual learning system that drastically optimizes system resources (lower memory, latency, energy consumption) while ensuring high accuracy.
Specifically, we (1) exploit meta-learning and rehearsal strategies to explicitly cope with data scarcity issues and ensure high accuracy, (2) effectively combine lossless and lossy compression to significantly reduce the resource requirements of CL and rehearsal samples, and (3) developed hardware-aware system on embedded and IoT platforms considering the hardware characteristics.

As a result, \sysname\ achieves near-optimal CL performance, falling short by only 2.8\% on accuracy compared to an Oracle baseline. With respect to the state-of-the-art (SOTA) Meta CL method, \sysname\ drastically reduces the memory footprint (by 178.7$\times$), end-to-end latency by 80.8-94.2\%, and energy consumption by 80.9-94.2\%.
In addition, we successfully deployed \sysname\ on two edge devices and a microcontroller unit, thereby enabling efficient CL on resource-constrained platforms where it would be impractical to run SOTA methods and the far-reaching deployment of adaptable CL in a ubiquitous manner. 
Code is available at \href{\repourl}{\color{magenta}{\repourl}}.

\end{abstract}



%
% Keywords. The author(s) should pick words that accurately describe
% the work being presented. Separate the keywords with commas.


\maketitle

\section{Introduction}\label{sec:introduction}

With the rise of embedded and Internet of Things (IoT) devices, the adoption of deep neural networks (DNN) has revolutionized various applications ranging from computer vision~\cite{he_deep_resnet}, audio~\cite{purwins_deep_2019} and sensing applications~\cite{lane_survey_2010}. However, in real-world setups, where a deployed model may need to dynamically learn new tasks (i.e., new classes or inputs) from users~\cite{chauhan_contauth_2020} and adapt to changing input distributions~\cite{pan_survey_2010}, existing learning approaches often fail, due to the constrained nature of available resources on edge devices and {\em catastrophic forgetting (CF)}~\cite{mccloskey_catastrophic_1989}.
CF describes the situation when a deployed model is able to perform new tasks but forgets previously learned knowledge. Efficient Continual Learning (CL) systems that can learn new tasks from growing data streams~\cite{parisi_continual_2019,chauhan_contauth_2020,Prabhu_2023_CVPR} are now being recognized as an important step forward as they also enable many practical applications. For example, household robotic devices need to continually learn to recognize new objects, while smart appliances need to learn different voice commands.

Many CL approaches have been proposed in the literature, including \textit{regularization-based}~\cite{kirkpatrick_overcoming_2017,zenke_continual_2017}, \textit{dynamic architecture-based}~\cite{rusu2016progressive,yoon_lifelong_2018,hung_compacting_2019}, and \textit{rehearsal-based methods}~\cite{rebuffi_icarl:_2017,pellegrini2019latent,chauhan_contauth_2020}. Among these, rehearsal-based methods largely alleviate the forgetting issue of a learned model. Nonetheless, they are excessively data-hungry as they require a large number of labeled samples to learn new information and to be stored as rehearsal samples~\cite{parisi_continual_2019}, incurring high computational and memory overheads.

\begin{figure}[t]
  \centering
  \subfloat[Performance]{
    \includegraphics[width=0.23\textwidth]{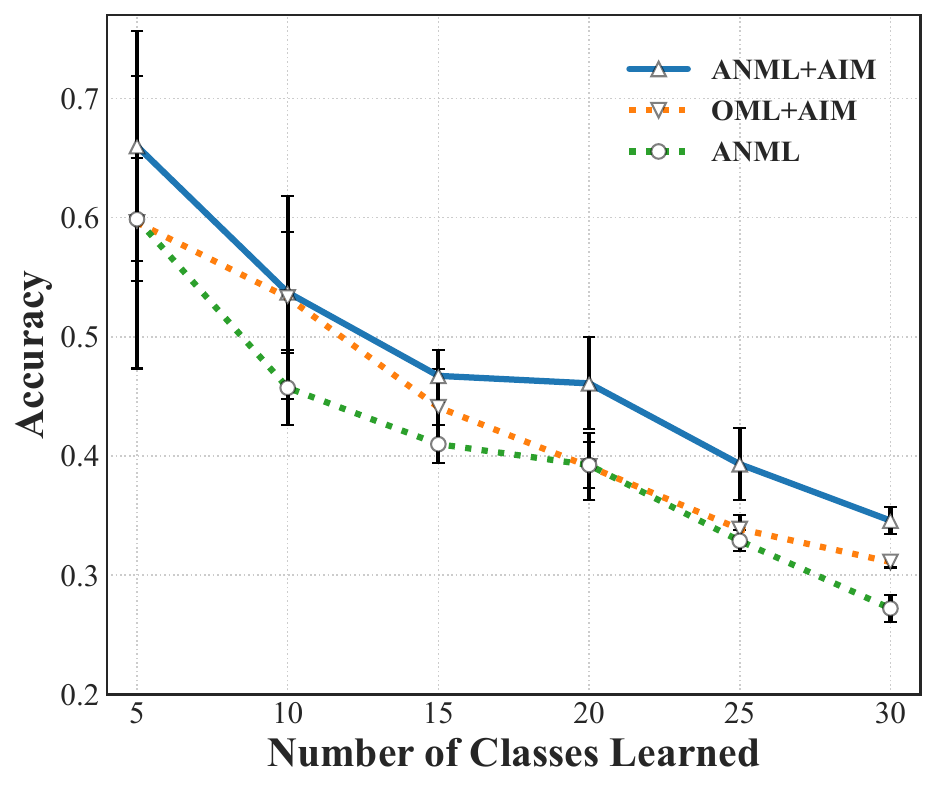}
  \label{subfig:accuracy}
  }
  \subfloat[Memory Overhead]{
    \includegraphics[width=0.23\textwidth]{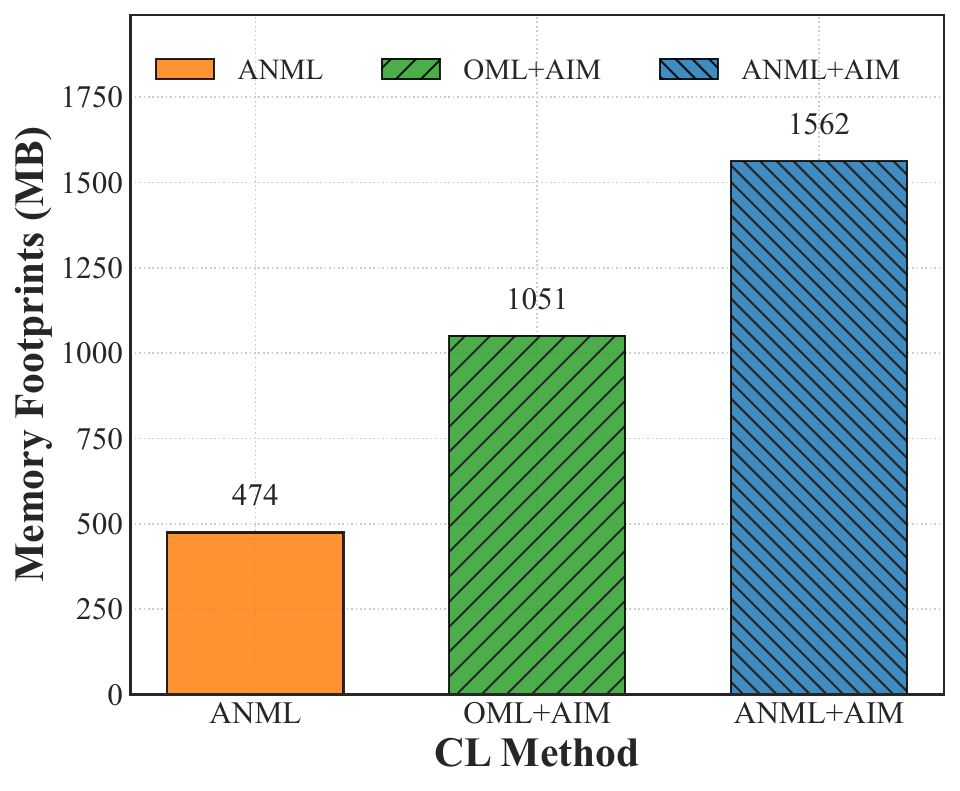}
  \label{subfig:modelsize}
  }
  \caption{
  Preliminary analysis of the prior Meta CL methods (i.e., ANML, OML+AIM, ANML+AIM). (a) shows the CL accuracy degradation of the Meta CL methods after learning $c$ number of classes on CIFAR-100~\cite{cifar10}. (b) shows the memory footprint needed to run the Meta CL methods on MiniImageNet~\cite{miniimagenet_vinyals_matching_2016} with a batch size of 8. 
  }
  \label{fig:main}
  \vspace{-0.1cm}
\end{figure}

Another stream of work has recently attempted to utilize meta-learning~\cite{hospedales_meta-learning-survey_2020} in CL to address the problem of the scarce labeled data.
A number of Meta CL methods~\cite{OML_javed_meta-learning_2019,ANML_beaulieu_learning_2020,AIM_lee_few-shot_2021} relying on a few samples of new classes to adapt and learn have been proposed. 
However, Meta CL’s performance degrades when many classes are added during deployment, leading to low scalability (refer to Figure~\ref{subfig:accuracy}). Additionally, state-of-the-art (SOTA) Meta CL methods, OML+AIM and ANML+AIM~\cite{AIM_lee_few-shot_2021}, exhibit large memory footprint, easily exceeding the RAM size on many embedded devices (e.g., 1 GB) (refer to Figure~\ref{subfig:modelsize}). 
Further, we observed that the end-to-end latency of SOTA Meta CL methods to continually learn multiple classes is computationally expensive. These aspects render prior Meta CL methods not deployable on resource-constrained devices. As such, there is an emerging need for novel system design approaches that facilitate the broader deployment of CL systems on various IoT devices by bringing down resource requirements of CL methods without jeopardizing their accuracy.

To address the aforementioned limitations, we develop \textit{\sysname}, the first hardware-aware system that fully enables data- and memory-efficient CL on the constrained edge and IoT devices. 
\textbf{First,} contrary to the existing Meta CL methods that primarily rely on regularization and suffer from accuracy loss, we introduce \textit{rehearsal-based Meta CL}; we co-design meta-learning with an efficient rehearsal strategy, enabling \sysname\ to rapidly learn new classes using only a few samples while alleviating catastrophic forgetting of the already learned classes upon deployment (Section~\ref{sub:rehearsal_metacl}).
\textbf{Second,} we propose a \textit{CL-tailored algorithm/software co-design approach} that minimizes the on-device resource overheads of CL. At the algorithmic level, we design a latent replay scheme, where rehearsal samples are extracted from an intermediate layer of the target DNN instead of holding copies of raw inputs. By strategically selecting the rehearsal layer for high compressibility, we facilitate the subsequent compression of rehearsal samples, enabling their efficient storage on-device. Besides, based on an observation that latent replays are sparse, we further design a novel \textit{Compression Module} via an intelligent combination of lossless compression to utilize sparsity and lossy compression to yield a high compression rate, fast encoding and decoding, and minimal resource usage (Section~\ref{sec:codesign}).
\textbf{Finally,} we develop our \textit{hardware-aware system} by employing hardware-friendly optimization techniques and considering the unique characteristics of hardware (e.g., write operation on Flash of IoT devices is costly during runtime) to optimize the runtime efficiency of CL operations on-device (Section~\ref{sec:implementation}).

\textbf{We make the following key contributions:}

\begin{enumerate}[leftmargin=*]
    \item A novel Meta CL method comprises a rehearsal strategy that alleviates catastrophic forgetting and a deployment-time inner-and outer-loop training structure that achieves both fast adaptation to new classes and refreshing of already learned classes. \sysname\ achieves previously unattainable levels of on-device accuracy, outperforming all existing Meta CL methods by 4.1-16.1\% on image and audio datasets, while being within 2.8\% of an oracle.

    \item A new algorithm/software co-design method that co-op\-ti\-mizes the rehearsal strategy and the compression pipeline to significantly reduce the resource requirements of CL and rehearsal samples. As a result, \sysname\ requires only 3.40--15.45~MB of memory and obtains a compression rate of 11.4--178.7$\times$ compared to the SOTA Meta CL method, ANML+AIM. This allows \sysname\ to run on edge devices, something impossible for current SOTA methods due to their large memory requirements ($>$1.05 GB).

    \item Our hardware-aware system implementation successfully deployed \sysname\ on two embedded devices (Jetson Nano and Raspberry Pi 3B+) and a microcontroller (STM32H747).
    Through extensive experiments, we demonstrate that \sysname\ outperforms existing CL and Meta CL baselines in terms of latency and energy consumption. Specifically, compared to ANML+AIM, \sysname\ obtains 80.8-94.2\% lower end-to-end latency and 80.9-94.2\% lower energy consumption on Jetson Nano. 
    Also, we developed \sysname\ on an extremely resource-constrained IoT device, STM32H747 with 512 KB of SRAM (2,000$\times$ smaller memory than Pi 3B+ with 1 GB RAM).
    To our knowledge, this is the first implementation of a CL framework onto this constrained and challenging platform, opening the door for the ubiquitous deployment of learning systems adaptive to users and environments over time continually.    
\end{enumerate}

\section{Background and Motivation}\label{sec:background}

\subsection{Continual Learning}\label{subsec:continual learning}

CL allows DNN models to learn over time from non-stationary data streams by acquiring new knowledge while avoiding the forgetting problem of the learned experiences~\cite{parisi_continual_2019,kwon_exploring_sec21,jha_continual_2021}.
In the literature, various approaches attempt to solve the forgetting problem~\cite{mccloskey_catastrophic_1989,mcclelland_why_1995,clsurvey_tpami_2021}.
The first group of approaches includes regularization-based methods~\cite{kirkpatrick_overcoming_2017,zenke_continual_2017,schwarz2018progress,aljundi_memory_2018,servia-rodriguez_knowing_2021}: these add a regularization term to the loss function to minimize changes to important weights of a model for previously learned classes to prevent forgetting. This approach can be very efficient regarding computation and memory costs. However, it is shown to be less effective than other methods that utilize additional resources such as expanding architectures and storing additional samples~\cite{clsurvey_tpami_2021}, as introduced in the following.
The second group of approaches includes the dynamic architecture-based methods~\cite{rusu2016progressive,yoon_lifelong_2018,hung_compacting_2019} that dynamically expand and freeze DNN architectures to incorporate new classes and prevent forgetting. Despite the promising performance, dynamic architectures pose the costly requirement of modifying the model architecture. This leads to higher computational costs as the model expands and prohibits the utilization of compile-time optimizations on a fixed computation graph of the model.
The last group of approaches among conventional CL includes rehearsal-based methods~\cite{rebuffi_icarl:_2017,lopez-paz_gradient_2017,hayes_remind_2020,pellegrini2019latent,chauhan_contauth_2020,kwon21_interspeech,Castro_2018_eie,Wu_2019_bic,Mittal_2021_essentials}. These prevent forgetting by replaying the saved rehearsal samples from earlier classes, typically leading to superior CL performance over the other methods at the cost of increased memory footprint.

In this work, we opt to use a rehearsal-based method due to its primarily superior performance in CL settings and the avoidance of dynamic expansion of the model architecture during deployment, allowing us to apply system optimizations on the static computation graph of the model (see last paragraph of Section~\ref{sec:implementation} for details).

Given a single trajectory of samples from a stream of classes $\mathcal{T}$, minimizing the CL loss of a DNN that is trained end-to-end is more challenging than conventional DNN training~\cite{OML_javed_meta-learning_2019}. This is because various complex challenges need to be solved together: (1) the forgetting problem incurred when learning a stream of different classes, (2) the issue with the lack of labeled samples, and (3) training DNNs is extremely sample-inefficient: the minimization problem requires multiple training epochs to converge to a reasonable solution.
Specifically, many CL methods~\cite{parisi_continual_2019,kwon_exploring_sec21} are proposed to alleviate the forgetting problem. However, they require a large amount of labeled data (a few thousand) and many training epochs. Another learning approach, called meta-learning, is proposed to make DNN more sample-efficient~\cite{hospedales_meta-learning-survey_2020,ding_rffsl_sensys20,lan_gazefsl_sensys20,xiao_onefi_sensys21}, requiring only a few samples to adapt/learn new data distributions from a correlated data stream~\cite{al-shedivat2018continuous,nagabandi2018deep}. However, existing meta-learning methods often neglect the forgetting problem of the already learned classes as it primarily aims at fast adaptation towards new tasks only~\cite{taesik_metasense_sensys19,finn_model-agnostic_2017,cdfsl,raghu_rapid_2019,triantafillou_meta-dataset_2020,snell_prototypical_2017,hu2022pmf,ondevicefsl_chauhan_2022}.

\subsection{Meta Continual Learning}\label{subsec:metacl}

To overcome the challenges mentioned thus far, researchers proposed a novel approach, Meta CL, that utilizes meta-learning in CL to enable data-efficient and fast adaptation to new classes and also attempts to alleviate forgetting of already learned classes through novel ways of regularization and/or modification of the model architecture~\cite{OML_javed_meta-learning_2019,ANML_beaulieu_learning_2020,AIM_lee_few-shot_2021}. First, to enable fast adaptation with only a few samples, Meta CL methods are based on the training procedure of meta-learning. The meta-learning uses an outer loop and an inner loop where the outer loop takes steps to improve the learning ability of the inner loop that optimizes the DNN model with a few samples. This phase is called \textit{meta-training}, which is typically performed on an offline server. The meta-training phase aims to find a better weight initialization of DNNs for fast adaptation with a few samples. After the meta-training is finished, the learned DNNs are tested given a few examples of new classes, referred to as the \textit{meta-testing} phase, that could run on embedded systems. Secondly, to prevent the forgetting problem, Meta CL methods separate the network architecture into the feature extractor and the classifier. During the meta-training phase, Meta CL adopts the concept of fast and slow learning on an architecture level. The feature extractor is updated in the outer loop (slow weights) using random samples from learned classes to prevent forgetting. The classifier is updated in the inner loop (fast weights) to learn new classes swiftly. This approach has proven useful in alleviating CF~\cite{OML_javed_meta-learning_2019,ANML_beaulieu_learning_2020,AIM_lee_few-shot_2021}.

\begin{figure*}[t]
  \centering
  \subfloat{
    \includegraphics[width=0.78\textwidth]{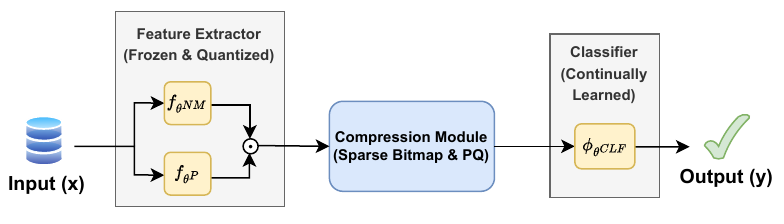}
  }
  \caption{
  The system overview. \sysname\ consists of the frozen/quantized feature extractor, the continually learned classifier, and the compression module based on sparse bitmap and PQ. The compression module takes the feature extractor's outputs (activations) as inputs and compresses them to be saved as latent replay samples.
  }
  \label{fig:overview}
  % \vspace{-0.2cm}
\end{figure*}

Although prior works in Meta CL enable CL with limited data samples, they have certain limitations. For example, Online-aware Meta-Learning (OML)~\cite{OML_javed_meta-learning_2019} and A Neuromodulated Meta-Learning (ANML)~\cite{ANML_beaulieu_learning_2020} can retain high CL performance on the Omniglot dataset~\cite{omniglot} over many classes. Also, Attentive Independent Mechanisms (AIM) module~\cite{AIM_lee_few-shot_2021} captures independent concepts to learn new knowledge. In fact, AIM and its combinations, ANML+AIM and OML+AIM, have achieved SOTA results. However, as prior Meta CL only relies on inner-loop optimization in the meta-testing phase, it does not utilize the concept of learning fast and slow weights during deployment.
Further, these methods fail to generalize (see Figure~\ref{subfig:accuracy}; low accuracy on CIFAR-100~\cite{cifar10}) and have extremely high memory requirements (see Figure~\ref{subfig:modelsize}), which limits their applicability to low-end devices. Hence, we aim to design an efficient Meta CL system that obtains high accuracy and less forgetting while making the practical deployment on embedded devices a reality.

\subsection{Efficient Deep Learning Systems}\label{subsec:codesign}

Scarce memory and compute resources are major bottlenecks in deploying DNNs on constrained embedded and IoT devices. In this context, researchers have largely focused on optimizing \textit{the inference stage} (i.e.,~forward pass) by proposing lightweight DNN architectures~\cite{gholami_squeezenext_2018,sandler_mobilenetv2_2018,ma_shufflenet_2018,lin_mcunet_2020,liberis_unas_2021}, pruning~\cite{han_deep_2016,liu_autocompress_2020}, quantization~\cite{jacob_quantization_2018,krishnamoorthi_quantizing_2018,rastegari_xnor-net_2016}, leveraging heterogeneous processors~\cite{jeong_band_mobisys22,neiwen_blastnet_sensys22,ling_rtmdl_sensys21}, and offloading computation~\cite{yao_deep_2020}.

In addition, many works focus on reducing the overall system resources required for \textit{DNN training}~\cite{cai_enable_2022,huang_elastictrainer_mobisys23,sohoni_low-memory_2019,pan2021mesa,evans_act_2021,chen_checkpoint_2016,jain_checkmate_2020,kirisame2021dynamic,huang_microbatching_2019,huang_swapadvisor_2020,kwon2023tinytrain,patil2022poet,wang_melon_2022,gim_memory-efficient_2022,tack2023learning,yue2021efficient,qu_pmeta_kdd2022}. For example, researchers control the layerwise growth of the model structure to enable efficient DNN training on mobile phones~\cite{zhang_mdldroidlite_2020}. Other methods optimize sparse activations and redundant weights to avoid unnecessary storage of activations and weight updates during DNN training~\cite{cai_tinytl_nips20,hosny_bittrain_2021,lin2022ondevice}. 
In particular, for memory-efficient training, researchers proposed efficient meta-learning approaches by tackling memory issues during meta-training~\cite{tack2023learning} and meta-testing~\cite{qu_pmeta_kdd2022}. 
However, dynamically changing the updated parameters as in~\cite{qu_pmeta_kdd2022} is not suitable to be used for MCUs because Flash memory space where the model weights are stored is read-only during runtime, and SRAM is even more limited than Flash in terms of memory capacity. Thus, it is difficult to incorporate the dynamic parameter update on MCUs.
Also, prior work~\cite{ko_lanecompress_tecs21} examines various lossless compression techniques (e.g., Huffman coding), which show at most a 3.3$\times$ compression ratio on activations. Lossy compression~\cite{chen2021actnn,liu_gact_2022} based on scalar quantization shows up to 12$\times$ memory savings without accuracy degradation. A promising method that can achieve even higher compression ratios (e.g.,~128$\times$) is Vector/Product Quantization (PQ)~\cite{jegou_product_2011,stock_and_2019,stock_training_2020}. However, as it requires storing a separate codebook containing representative vectors, a brute-force utilization of PQ may not achieve actual memory savings. In this work, we demonstrate that PQ can be a key component towards efficient continuous learning and show how the on-device CL pipeline should be designed to accommodate it (see Section~\ref{sec:compr_module} and Figure~\ref{fig:compression_module} for details).

In contrast to previous works, \sysname\ realizes efficient continual learning that was previously considered impractical for many embedded devices. By developing rehearsal-based Meta CL, effective algorithm/software co-design, and hardware-aware system implementation considering the unique characteristics of a wide range of embedded and IoT platforms (e.g., Jetson Nano, Pi 3B+, and STM32H747), \sysname\ yields both high accuracy and low resource overheads.

\vspace{-0.2cm}

\section{L\MakeLowercase{ife}L\MakeLowercase{earner}}\label{sec:methodology}

\sysname\ leverages the idea of Meta CL and rehearsal-based learning and minimizes the system overheads on embedded devices. \sysname\ consists of two phases. The first phase, i.e., meta-training, is performed on a server to obtain a good weight initialization by utilizing meta-learning in the CL setup with a few samples. The second phase is meta-testing: a meta-trained model is deployed on embedded devices and learns new classes continually without forgetting previously learned classes.
Additionally, as shown in Figure~\ref{fig:overview}, \sysname\ has two components to ensure superior performance and efficiency when it is deployed on resource-constrained devices: (1)~co-utilization of Meta CL and rehearsal strategy together with a deployment-time inner- and outer-loop optimization to resolve the accuracy degradation issue, (2) a design scheme that co-optimizes \sysname's rehearsal strategy and compression pipeline (\textit{Compression Module} in Figure~\ref{fig:overview}) 
to minimize the memory footprint, compute cost, and energy consumption when running CL.

\subsection{Co-utilization of Meta-Learning and Rehearsal Strategy}\label{sub:rehearsal_metacl}  

\begin{figure*}[t]
  \centering
  \subfloat{
    \includegraphics[width=0.8\textwidth]{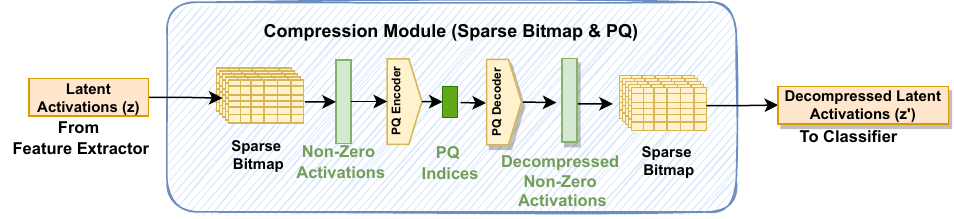}
  }
  \caption{
  The overview of our compression module. It consists of (1) a sparse bitmap to filter out zero from activations or to reconstruct decompressed activations from non-zero activations, (2) a PQ encoder that further compresses non-zero activations into PQ indices, and (3) a PQ decoder that decompresses PQ indices back into decompressed non-zero activations.
  }
  \label{fig:compression_module}
  % \vspace{-0.2cm}
\end{figure*}

Current Meta CL methods rely on regularization in order to minimize radical changes to the already trained weights when learning new classes. As such, given a small set of training data from a stream of classes, all samples are discarded once they have been used. 
However, recent results from the CL literature~\cite{clsurvey_tpami_2021} indicate that the alternative approach of rehearsal-based methods often outperforms regularization-based CL. Driven by this observation, we design our Meta CL method, called \textit{rehearsal-based Meta CL}, which introduces a rehearsal strategy into the Meta CL to improve CL performance. Concretely, we introduce a Replay Buffer that stores informative samples from already learned classes; these serve as additional training samples when learning new classes, form a mechanism for refreshing the weights of the model, and avoid catastrophic forgetting.

In addition, existing Meta CL systems are limited by their sole use of inner-loop optimization during meta-testing. 
Instead, we construct a variant of the learning fast and slow weights approach: we utilize the samples of new classes during inner-loop updates to enable rapid adaptation to new classes, followed by outer-loop iterations with the rehearsal samples of the previously learned classes to alleviate catastrophic forgetting.

\textbf{System Overhead.}~Despite the learning benefits of our rehearsal-based Meta CL method (see Section~\ref{subsec:performance} for details), it comes at a system cost. With respect to memory, the Replay Buffer has to store a number of representative samples for each of the already encountered classes, so that they can be fetched during meta-testing. With respect to computation, the samples have to be processed by the DNN with both forward and backward passes to perform CL. Unless alleviated, these overheads can lead to a sharp increase in storage and computational requirements, hindering its deployment on mobile and embedded devices, where continual learning is most needed. In the next section, we present \sysname's co-design approach for alleviating these system costs.

\subsection{CL-tailored Algorithm/Software Co-Design}\label{sec:codesign}
To alleviate the system costs of rehearsal-based Meta CL and enable its deployment on resource-constrained devices, we present an algorithm-software co-design method, optimized for Continual Learning. At the algorithmic level, we design a \textit{rehearsal strategy} that minimizes the computational overhead while maximizing the compressibility of the rehearsal samples. At the software level, we design a two-stage \textit{Compression Module} that enables the efficient compression, storage and decompression of rehearsal samples, while inducing minimal on-device resource usage.

\subsubsection{Rehearsal Strategy.}
Key design decision in rehearsal-based methods constitutes the form of the rehearsal samples. A standard approach followed by many CL methods~\cite{rebuffi_icarl:_2017,lopez-paz_gradient_2017,chauhan_contauth_2020} is \textit{native rehearsal (i.e.,~raw data replay)}, which stores and replays the input data in their raw format, e.g.,~images are stored for computer vision tasks and MFCC features for audio tasks. Under this scheme, a random subset of the given classes is stored as rehearsal samples, which are later replayed to mitigate the forgetting issue. The drawbacks of this approach are the significant computational overhead, as the samples have to be processed from the full model, and the compression variability as compressibility varies substantially in a per-sample manner.

To counteract these drawbacks, we introduce \textit{latent replay} into our rehearsal strategy. Under this scheme, instead of holding copies of raw inputs, we store their latent representations, i.e.,~intermediate activations at the output of a selected layer of the target DNN. In \sysname, we employ two techniques in order to enable the utilization of latent replay: \textit{i)}~select the last layer of the model's feature extractor as the rehearsal point; and \textit{ii)}~we freeze the feature extractor upon deployment and perform CL only on the classifier. With the feature extractor frozen, we render latent replay functionally equivalent to raw data replay. On the computational front, the forward pass of the feature extractor can be omitted when replaying latent representations and the backward propagation is performed until the last layer, inducing significant computational gains. 

On the memory front, we make the following observation.
In DNN training, the activations for each layer are saved during the forward propagation so that those activations are utilized for computing the gradients during the backward propagation. As in~\cite{sohoni_low-memory_2019}, storing activations requires a large memory footprint depending on the batch size used for training. However, commonly used ReLU non-linearity in many DNN models results in sparse activations in the successive layers. Also, we observe that more than 90\% of the activation values of the latent layer are zero due to the usage of ReLU from our analysis of the network architecture on all three datasets. By strategically selecting the rehearsal layer in the DNN and treating ReLU activations as the rehearsal samples, \sysname's rehearsal strategy facilitates their compression and subsequent efficient storage on-device.

\subsubsection{Compression Module for Latent Replays}
\label{sec:compr_module}

We now introduce the \textit{Compression Module} that is responsible for \textit{i)}~compressing rehearsal samples (i.e.,~latent activations in our work) when new classes are encountered and storing them in the Replay Buffer, and \textit{ii)}~fetching and decompressing them to perform CL at runtime. This component comprises two stages: sparse bitmap compression and product quantization (PQ).

\textbf{Sparse Bitmap Compression.}
To leverage the sparsity of our latent replays for efficient storage, we employ sparse bitmap compression~\cite{hosny_bittrain_2021}. This scheme enables the Compression Module in \sysname\ to filter out the majority of zero values (typically 90\% or more) in latent activations and save the remaining non-zero values to increase the compression rate for saving latent activations. 

Figure~\ref{fig:compression_module} depicts the compression and decompression processes. For compression, when latent activations are given to our system, a bitmap with the same dimensions as the latent activations sets a bit to 1 for non-zero values' indices and 0 for the remainders. Then, non-zero values and the sparse bitmap are stored in 32-bit floats and the bitmap format, respectively.
For decompression, we traverse all elements of the bitmap and a vector containing the stored non-zero values, reconstructing in this process the latent activations by using either the saved non-zero value or zero if a bitmap element is 1 or 0, respectively. 
The compression and decompression processes are linear in runtime: $O(n)$, where $n$ is the total number of elements of latent activations. With respect to memory, the footprint is reduced from $(4n)$ when a dense format is used for storing latent activations to $(4\times\textnormal{number of non-zero values} + \frac{1}{8}n)$ with the bitmap.

\textbf{Product Quantization.}\label{subsec:pq}
To further minimize the resource overhead of rehearsal samples, we introduce a second stage to our compressor (Figure~\ref{fig:compression_module}) utilizing PQ~\cite{jegou_product_2011}. The output of the sparse bitmap compressor contains a vector of non-zero values.
With PQ being a vector compression method that can compress a given vector $\mathbf{v} \in \mathbb{R}^d$ into $s$ number of PQ indices using a PQ codebook with $s$ columns, it is suitable to further reduce the size of the encoded rehearsal samples. Each column of the PQ codebook contains a set of representative vectors that well approximate $s$ sub-vectors of $\mathbf{v}$ when $\mathbf{v}$ is partitioned into $s$ sub-vectors.

For compression, the PQ encoder applies PQ to the non-zero activations $\mathbf{v} \in \mathbb{R}^d$ that are already filtered out by the first-stage sparse bitmap compression.
We use 1 byte to store each PQ index and set $d/s = \{128, 32, 8\}$ (length of each sub-vector). Then, each sub-vector of length $d/s$ containing 32-bit floats is encoded to a 1-byte PQ index via our PQ encoder for more analysis regarding hyper-parameters).
\sysname\ learns the PQ codebook offline using the latent activations during the meta-training phase, which is then stored on-device.
For decompression, the PQ decoder reconstructs the non-zero activations $\mathbf{v'}$ using the stored PQ indices and the PQ codebook.

Finally, as in Algorithm~\ref{alg:meta-test} (see Lines 7, 9, and 10), our compression module is seamlessly incorporated in the inner- and outer-loop optimization of \sysname, enabling on-the-fly compression of the latent activations during deployment.

\subsection{Putting It All Together}

Having described the main components of \sysname\, we now present the complete meta-training and meta-testing procedures that take place offline and online, respectively.

\begin{algorithm}[!t]
\caption{Meta-Training Procedure of \sysname}
\label{alg:meta-train}
\SetAlgoNoLine
\DontPrintSemicolon
  \KwRequire{$N$ sequential classes $\mathcal{T}$; learning rates (LR) $\alpha$, $\beta$; inner-loop iterations $k$; modules $f_{\theta}, \phi_{\theta}$; given samples $S$
  }
    
    \SetAlgoVlined
    \tcp{Outer-loop starts here}
    \For{$t = 1, ..., N$} 
    {
        $S_{traj}\sim \mathcal{T}_t,\ S_{rand}\sim \mathcal{T}$\;
        
        \tcp{Inner-loop starts here}
        \For{$i = 1, ..., k$}
        {
            Update fast weights using $S_{traj} \qquad \quad \rhd $ LR: $\alpha$\;
            
            \tcc{OML(+AIM):$\phi_{\theta^{PLN}} (f_{\theta^{W}})$, ANML(+AIM):$f_{\theta^{P}}, \phi_{\theta^{CLF}} (f_{\theta^{W}})$,
            \sysname:$\phi_{\theta^{CLF}}$}
        }
        Update slow weights using $\{S_{traj},S_{rand}\}$ $\rhd$ LR: $\beta$\;
        
        \tcc{OML(+AIM):$f_{\theta^{RLN}}$,\ 
        ANML(+AIM):$f_{\theta^{NM}}, f_{\theta^{P}}, \phi_{\theta^{CLF}}$,
        \sysname:$f_{\theta^{NM}}, f_{\theta^{P}}, \phi_{\theta^{CLF}}$}
    }
\end{algorithm}

\begin{algorithm}[!t]
\caption{Meta-Testing Procedure of \sysname}
\label{alg:meta-test}
\SetAlgoNoLine
\DontPrintSemicolon
  \KwRequire{$N$ sequential unseen classes $\mathcal{T}$; learning rates (LR) $\alpha$, $\beta$; inner-loop iterations $k$; modules $f_{\theta}, \phi_{\theta}$, $BitPQ_{compress, decompress}$; samples $S$ \setcounter{AlgoLine}{0}
  }
    
    \SetAlgoVlined
    $S_{train} = \{\},\ S_{rehearsal} = \{\}$\;
    
    \tcp{Outer-loop starts here}
    \For{$t = 1, ..., N$} 
    {
        $S_{traj}\sim \mathcal{T}_t$\; 
        
        $S_{train} = \{S_{train}, S_{traj}\}$\;
        
        \tcp{Inner-loop starts here}
        \For{$i = 1, ..., k$}
        {
            Update fast weights using $S_{traj} \qquad \quad \rhd $ LR: $\alpha$\;
            
            \tcc{OML(+AIM):$\phi_{\theta^{PLN}} (f_{\theta^{W}})$, ANML(+AIM):$f_{\theta^{P}}, \phi_{\theta^{CLF}} (f_{\theta^{W}})$,
            \sysname:$\phi_{\theta^{CLF}}$}
        }
        \tcp{Get latent activations from compressed rehearsal samples}
        $S_{latent} = BitPQ_{decompress}(S_{rehearsal}) $\; 
        
        Update slow weights using $\{S_{traj},S_{latent}\} \ \rhd$ LR: $\beta$\;
        
        \tcc{OML(+AIM):$f_{\theta^{RLN}}$,\ ANML(+AIM):$f_{\theta^{NM}}, f_{\theta^{P}}, \phi_{\theta^{CLF}}$,
            \sysname:$\phi_{\theta^{CLF}}$
            }
        
        \tcp{Get latent activations}
        $S_{latent} = f_{\theta^{NM}} (S_{traj}) \odot f_{\theta^{P}}(S_{traj}) $\; 
        
        \tcp{Store compressed activations for rehearsal}
        $S_{rehearsal} = \{S_{rehearsal},\ BitPQ_{compress}(S_{latent}) \}$\;
    }
    $S_{test} = \mathcal{T} - S_{train}$ \tcp*{Held-out test set}
    Evaluate on $S_{train},\ S_{test}$ \tcp*{Eval on training/test set}    
\end{algorithm}

\textbf{Meta-Training Procedure.}
Algorithm~\ref{alg:meta-train} shows the procedure of meta-training of Rehearsal-based Meta CL, \sysname. Firstly, the meta-training process of rehearsal-based Meta CL is the same as that of Meta CL~\cite{ANML_beaulieu_learning_2020}. In detail, it is comprised of an inner loop inside an outer loop of optimization. In the inner loop, the classifier part is updated (fast weights, e.g., $\theta^{PLN}$ for OML and $\theta^{P,CLF}$ for ANML, $\theta^{PLN,W}$ for OML+AIM, and $\theta^{P,CLF,W}$ for ANML+AIM) (Lines 4-5). The number of weight update iterations is determined by the number of samples $k$ (e.g., 10-30) of a given sample set, $S_{traj}$, of a new class, $\mathcal{T}_t$. After the $k$ sequential updates, the meta-loss in the outer loop (Line 6) is computed using all the given samples on the new class ($S_{traj}$) and randomly sampled samples from all the meta-training classes ($S_{rand}$).
All the weights of DNN are updated through outer-loop gradient updates using an Adam optimizer~\cite{kingma_adam_2017}. The learning rates, $\alpha$ for the inner loop and $\beta$ for the outer loop, are used as hyper-parameters.

\textbf{Meta-Testing Procedure.}
After executing the meta-training phase on a server, our system is deployed on resource-constrained devices and evaluated on its ability to learn unseen classes in the meta-testing phase. Algorithm~\ref{alg:meta-test} shows the meta-testing phase of the rehearsal-based Meta CL.
In prior Meta CL, the meta-testing procedure contains only inner-loop optimization without outer-loop optimization, i.e., only fast weights except for slow weights are fine-tuned.
In contrast, \sysname\ leverages the full potential of meta-learning by using both inner-and outer-loop optimization in the meta-testing phase. Specifically, our proposed meta-testing procedure starts with the inner-loop weight updates to learn new classes swiftly using a few samples (Lines 5-6), followed by the outer-loop weight updates to retain the knowledge on the previously learned classes using the replayed samples plus the new samples (Line 8). Note that although the outer-loop iteration could run multiple epochs, the performance converges after one or two epochs (refer to Section~\ref{subsec:parameter analysis} for more analysis). Also, \sysname\ integrates the compression module that compresses (Lines 9-10) and decompresses (Line 7) the latent activations during outer-loop optimization, as described in Section~\ref{sec:codesign}.

\textbf{Our Contribution.}
Our method conceptually leverages existing concepts. We solve the challenge of incorporating these concepts in a coordinated, efficient end-to-end system 
(as discussed in Section~\ref{subsec:codesign}). We achieve higher accuracy than baselines while reducing the memory footprint drastically. Our key contributions are (1) co-designing the algorithmic innovation (rehearsal strategy) with an intelligent combination of lossless (bitmap) and lossy (PQ) compression to significantly reduce the resource requirements of CL and latent replay samples (Section~\ref{sec:methodology}), (2) successfully deploying \sysname\ end-to-end on two embedded devices and MCU on which many prior works fail to run (Section~\ref{sec:implementation}).

\vspace{-0.2cm}
\section{Hardware-Aware System Implementation}\label{sec:implementation}
We develop the first phase, meta-training, of Meta CL methods on a Linux server to initialize the neural weights that can enable fast adaptation during deployment scenarios. 
After that, for the second phase, meta-testing, (i.e., actual deployment scenarios), we implemented our hardware-aware system by considering the hardware capacity and unique runtime characteristics of our target devices: (1) embedded and mobile systems such as Jetson Nano and Raspberry Pi 3B+, and (2) a microcontroller unit such as STM32H747. To further optimize the system efficiency, we adopt hardware-friendly optimization techniques in our implementation\footnote{\href{\repourl}{\color{magenta}{\repourl}}}

\textbf{Embedded Device.} 
Jetson Nano has a quad-core ARM Cortex-A57 processor, and 4 GB of RAM, while Pi 3B+ contains a quad-core ARM Cortex-A53 processor with 1 GB of RAM. Note that the free memory space of Jetson Nano and Pi 3B+ during idle time is roughly 1.7 GB and 600 MB, respectively, due to the memory footprints pre-occupied by background, concurrent applications, and an operating system. 
As software platforms, we employ Faiss (PQ Framework)~\cite{johnson_billion-scale_2019} and PyTorch 1.8 (Deep Learning Framework)~\cite{pytorch} to develop and evaluate the meta-training and meta-testing phases on embedded systems. 

\textbf{Microcontroller Unit (MCU).}
To demonstrate the feasibility of the broader deployment of CL systems at the extreme edge, we further optimized and developed \sysname\ on MCUs. We implemented the online component of \sysname\ using C++ on an STM32H747 device equipped with ARM Cortex M4 and M7 cores with 1MB SRAM and 2 MB eFlash in total. However, we only utilize one core (ARM Cortex M7), as most MCUs have one CPU core. Also, we restrict the usage space of SRAM and eFlash to 512 KB and 1 MB, respectively, to enforce stricter resource constraints (an order of magnitude smaller memory space than other embedded devices with larger than 1 GB RAM).

To deploy \sysname\ on MCUs effectively and efficiently, we addressed many technical challenges and considered hardware characteristics. First of all, the memory requirements of the MetaCL methods developed on embedded devices, including \sysname, far exceed the hardware capacity of a "high-end" MCU such as STM32H747 (refer to Section~\ref{subsec:performance}). Hence, we first searched for a smaller yet accurate architecture for MCUs by experimenting with various width modifiers~\cite{sandler_mobilenetv2_2018,lin_mcunet_2020,liberis_unas_2021} (see Section~\ref{subsec:mcu deploy} for details).

We then implemented our Compression Module (sparse bitmap compression and PQ) to reduce memory usage of latent replay samples on SRAM.
In particular, we consider hardware characteristics and constraints: (1) the write operation on the storage (Flash) of MCUs is costly~\cite{svoboda_mcudeploy_euromlsys22}, and (2) Flash is read-only during runtime~\cite{banbury_micronets_2021,kwon_yono_ipsn22}. Hence, in our MCU implementation of \sysname, to minimize the memory footprint and energy consumption required for latent replay, we first compress latent replay samples using our Compression Module and then store them on SRAM, which has more limited memory but is faster and cheaper to perform read/write operations on than Flash. Note that our learned PQ codebook, used to encode and decode the latent replay samples after sparse bitmap compression, is stored on Flash to leave more space for scarce resources of SRAM. Also, PQ codebooks are static once deployed; they can be stored on the read-only memory of Flash.

In addition, we rely on the TFLM framework~\cite{robert_tflm_mlsys2021} to perform inference of the feature extractor on MCUs. However, TFLM does not support training (i.e., backpropagation). We developed our Backpropagation Engine based on C/C++ using Eigen~\cite{eigenweb} as a data structure and matrix multiplication library. Based on our Backpropagation Engine, we construct the classifier part on the fly whose weights are allocated on SRAM and can be continually learned during deployment whenever more data for new classes become available. Our lightweight Backpropagation Engine enables the implementation of the first CL system on MCUs.

Lastly, the binary size of our Compression Module and Backpropagation Engine, excluding C++ Standard Library (STL) on an MCU, is only 80 KB, introducing minimal overhead on storage.

\textbf{Hardware-friendly Optimization.}
We further optimize \sysname's CL operations on-device. By freezing the model's feature extractor during deployment, \sysname\ significantly reduces the computational cost for the already learned classes during replay by omitting the forward and backward passes. In addition, we utilize the hardware-friendly 8-bit integer arithmetic~\cite{sze_efficient_2017} by reducing the precision of weights/activations of the feature extractor from 32-bit floats to 8-bit integers, increasing the computation throughput and minimizing latency and energy. The scalar quantization scheme~\cite{jacob_quantization_2018,krishnamoorthi_quantizing_2018} is used to minimize the information loss in quantization. Then, we utilize the QNNPACK~\cite{qnnpack} backend engine and TFLM to execute the quantized model on two embedded devices and MCUs, respectively.

\section{Evaluation}\label{sec:evaluation}

\subsection{Experimental Setup}\label{subsec:setup}
We briefly describe our experimental setup in this subsection.

\subsubsection{Metrics} 
As in~\cite{ANML_beaulieu_learning_2020}, we use testing accuracy on unseen samples of all the new classes learned continually as a key performance metric, representing the generalization ability of CL systems. In addition, we measure the memory footprint (model parameters, optimizers, activations, and rehearsal samples), end-to-end training latency and energy consumption to continually learn all the given classes for a deployed DNN on embedded devices.

\subsubsection{Datasets}

We employ three datasets of two different data modalities in our evaluation.

\textbf{CIFAR-100~\cite{cifar10}:} 
Following~\cite{AIM_lee_few-shot_2021}, we employ CIFAR-100 in our evaluation as it is widely used dataset. CIFAR-100 consists of 60,000 images of 100 classes. Each class has 500 train images and 100 test images. 70 classes are used for meta-training and the remaining 30 for meta-testing. During both meta-training and meta-testing, up to only 30 training images are sampled for training in each class, which holds for both MiniImageNet and GSCv2 datasets. Then, during meta-testing, a total of 900 samples are given to perform~CL.

\textbf{MiniImageNet~\cite{miniimagenet_vinyals_matching_2016}:}
Following~\cite{AIM_lee_few-shot_2021}, we employ MiniImageNet containing 64 classes for meta-training and 20 classes for meta-testing. Each class has 540 images for training and 60 images for testing. During meta-testing, a total of 600 samples are given.

\textbf{GSCv2~\cite{gscv2}:} 
To generalize our results to another data modality, we include Google Speech Command V2 (GSCv2) as it is a widely used audio dataset. GSCv2 consists of a total of 35 classes of different keywords. We use 25 classes for meta-training and 10 classes for meta-testing. Each class has 2,424 and 314 input data for training and testing, respectively. During meta-testing, 300 samples in total are given for CL.

\begin{figure*}[t!]
  \centering
  \subfloat[CIFAR-100]{
    \includegraphics[width=0.325\textwidth]{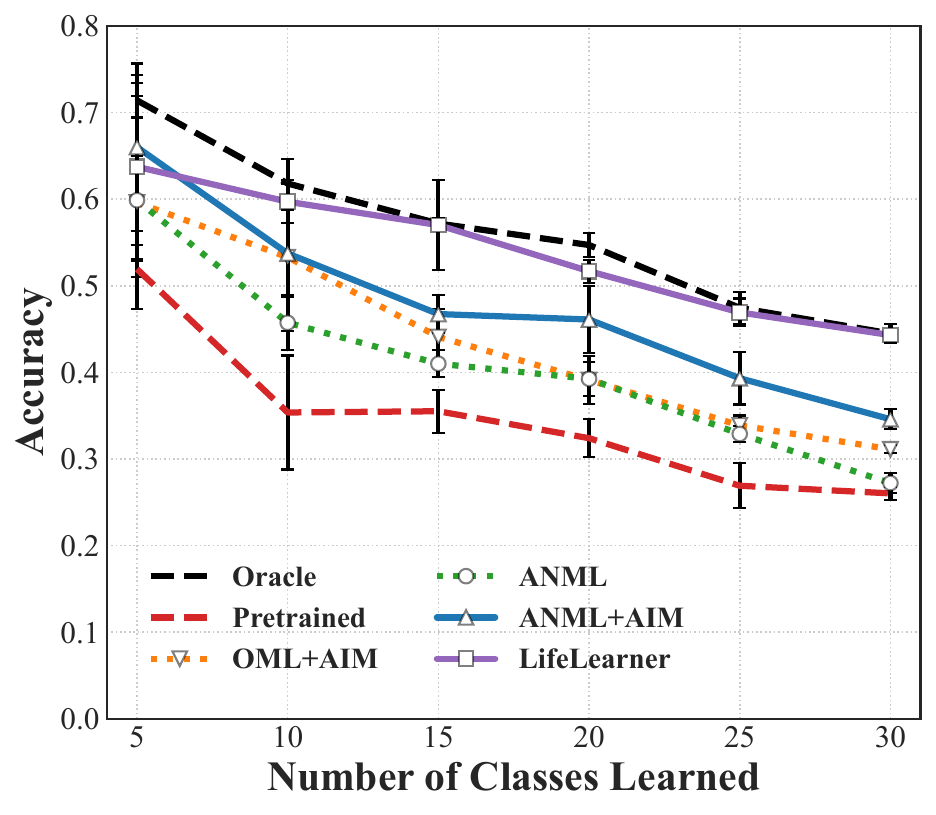}
  \label{subfig:cifar_test}}
  \subfloat[MiniImageNet]{
    \includegraphics[width=0.325\textwidth]{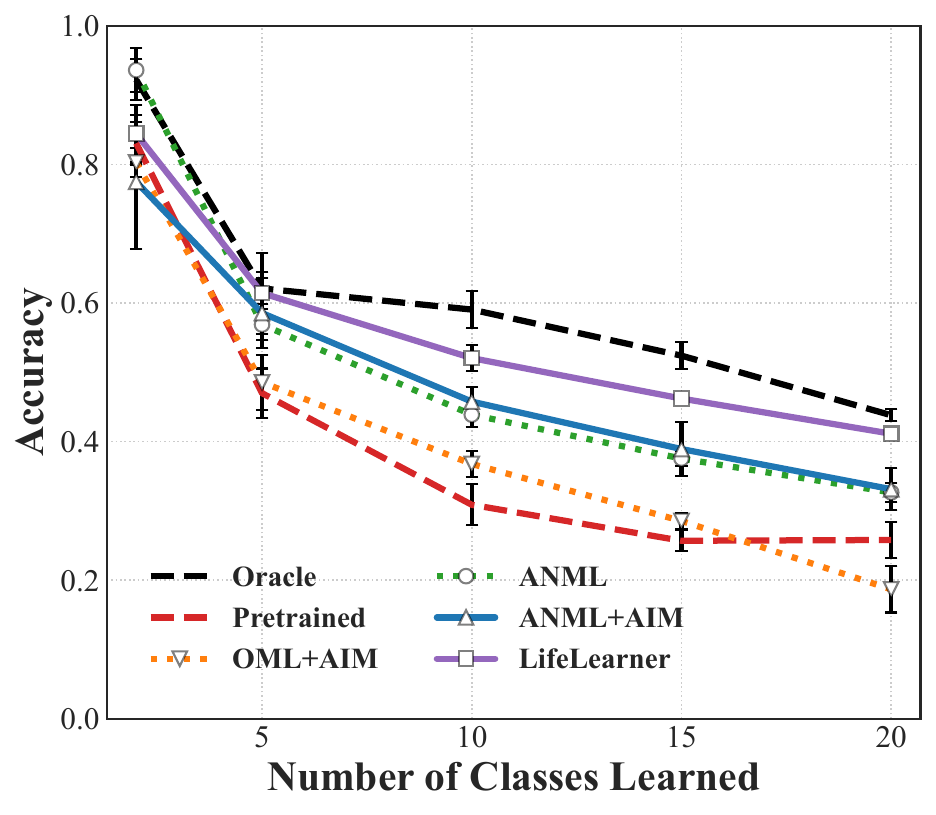}
  \label{subfig:miniimagenet_test}}
  \subfloat[GSCv2]{
    \includegraphics[width=0.325\textwidth]{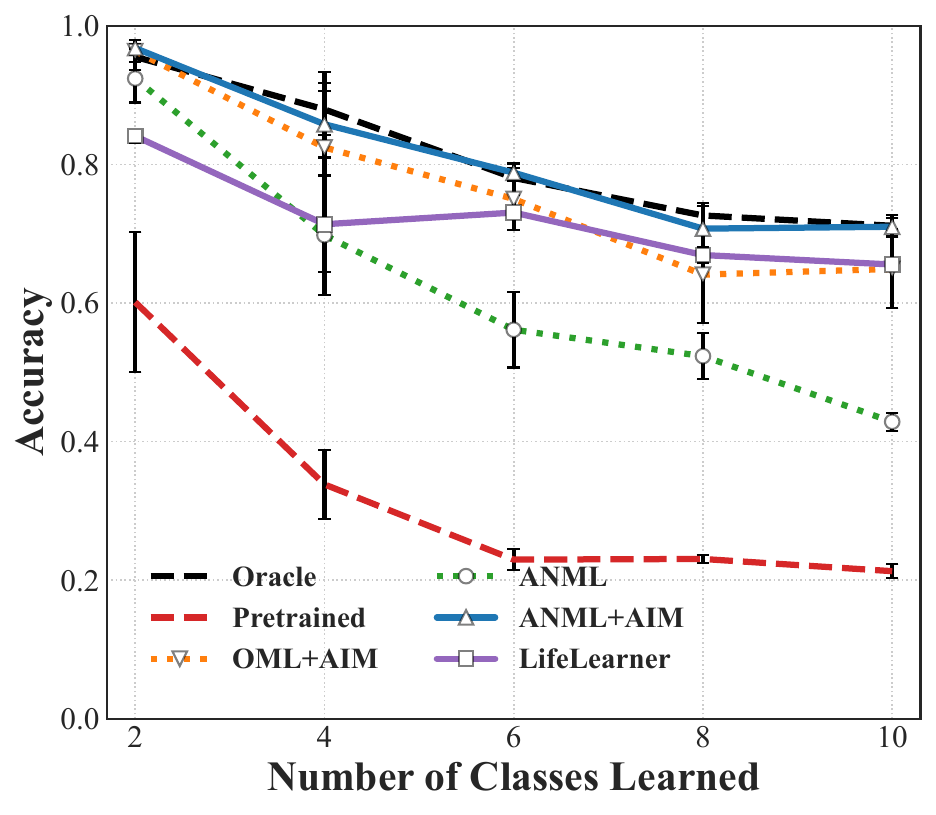}
  \label{subfig:gscv2_test}}
  \caption{
  The accuracy of the CL systems on the three datasets of two different modalities. Reported results are averaged over three trials, and standard-deviation intervals are depicted.
  }
  \label{fig:performance}
  % \vspace{-0.2cm}
\end{figure*}

\subsubsection{Baselines}
We compare our system, \textit{\sysname}, with five baseline systems as follows.

\textbf{Oracle:} The CL performance of Oracle represents the upper bound performance of the experiments. It is because Oracle has access to all the classes at once in an i.i.d. fashion and performs DNN training for many epochs until the performance converges.

\textbf{Pretrained:} This baseline initializes the model weights based on conventional DNN training without the meta-learning procedure. Then, it finetunes the weights using given samples in the meta-test phase, similar to prior Meta CL methods.

\textbf{OML+AIM~\cite{AIM_lee_few-shot_2021}:} This is a Meta CL method based on OML with an Attentive Independent Mechanisms (AIM) module, capturing independent concepts to learn new knowledge.

\textbf{ANML~\cite{ANML_beaulieu_learning_2020}:} It is the representative Meta CL method. As this method is often reported to outperform OML~\cite{OML_javed_meta-learning_2019}, we only employ ANML in our evaluation. Also, note that the proposed components of \sysname\ build on top of ANML.

\textbf{ANML+AIM~\cite{AIM_lee_few-shot_2021}:} ANML+AIM is a Meta CL method based on ANML with an AIM module. This baseline serves as the SOTA Meta CL method as it often outperforms other Meta CL methods including OML+AIM.

\subsubsection{Model Architecture}
\sysname\ employs the network architecture used in the prior CL works for a fair comparison~\cite{ANML_beaulieu_learning_2020,AIM_lee_few-shot_2021}. As in Figure~\ref{fig:overview}, it consists of the feature extractor and the final classifier. 
For ANML-based model architectures, the feature extractor consists of a neuromodulatory network, $f_{\theta^{NM}}$, and a prediction network, $f_{\theta^{P}}$, followed by the classifier part, $f_{\theta^{CLF}}$. 
The neuromodulatory and prediction networks are 3-layer convolutional networks with 112 and 256 channels, respectively. The classifier has a single fully-connected layer. In this case, \sysname\ utilizes the last layer of the feature extractor as the latent replay layer, following the natural structure of the ANML architecture.\footnote{When targeting a different model architecture, the latent replay layer selection is a configurable design decision. We leave this investigation as future work.}
The SOTA method, ANML+AIM, adds AIM layers $f_{\theta^{W}}$ between the feature extractor and the classifier, which alleviates forgetting and helps learn new classes. 
In addition, for OML and OML+AIM, the feature extractor has a 6-layer convolutional network with 112 channels, followed by the classifier of two fully-connected layers with an AIM module between the feature extractor and the classifier.
Note that the model architectures deployed on embedded devices (i.e., Jetson Nano and Pi 3B+) and an MCU (i.e., STM32H747) are different due to the strict resource constraint on the MCU. Thus, a smaller version of the model architecture described above is adopted for the MCU deployment (see Section~\ref{subsec:mcu deploy} for details).

\subsubsection{Training Details}
We followed the meta-training procedure used in prior Meta CL works~\cite{OML_javed_meta-learning_2019,ANML_beaulieu_learning_2020,AIM_lee_few-shot_2021}. For instance, we used a batch size of 1 and 64 for the inner- and outer-loop updates over 20,000 steps, respectively. We experimented with different learning rates for the inner loop and outer loop to obtain the meta-trained DNN that provides the best accuracy on a validation set. 
As a result, for CIFAR-100 and GSCv2 datasets, the inner-loop learning rate ($\alpha$) is set to 0.001, and the outer-loop learning rate ($\beta$) is also set to 0.001. For the MiniImageNet dataset, the optimal settings are $\alpha$ = 0.001 and $\beta$ = 0.0005.
During the meta-testing phase, ten different learning rates are tried for all the methods, and the best-performing results are reported.
Besides, to obtain the accuracy results of systems that perform replays, we experimented with batch sizes of 8 and 16 and observed little difference in CL performance. Thus, we employ a batch size of 8, as a smaller batch size reduces the memory footprint.

\subsection{Experimental Results}\label{subsec:performance}

\textbf{Accuracy.}
We start by evaluating the CL performance (testing accuracy) of \sysname\ compared to the baselines on the employed datasets. 
Figure~\ref{fig:performance} presents the accuracy results of the meta-testing phase. Pretrained serves as the lower bound. The low accuracy (24.4\% on average for three datasets) of Pretrained demonstrates that the conventional transfer learning approach cannot address the challenging scenarios of learning new classes with only a few samples. ANML improves upon Pretrained, however, the improvement is marginal (i.e., average 9.9\% accuracy gain compared to Pretrained but 18.9\% accuracy drop on average compared to Oracle which shows the upper bound accuracy). 
Note that it is very challenging to achieve high testing accuracy even for Oracle as the number of available samples is very limited during meta-testing: all evaluated systems are given only 30 samples per class, accounting for only 2.57\%, 1.74\%, and 0.5\% of all training samples during meta-training of CIFAR-100, MiniImageNet, and GSCv2, respectively.

\sysname\ achieves near-optimal CL performance, falling short by only 2.8\% accuracy compared to Oracle. Also, \sysname\ outperforms all the Meta CL methods with substantial accuracy gains of 4.1-16.1\% on average for the three datasets. Specifically, \sysname\ shows almost no loss of accuracy, i.e., 0.2\% for CIFAR-100 and 2.7\% for MiniImageNet compared to Oracle. In contrast, ANML+AIM (i.e., the previous SOTA Meta CL method) shows notable accuracy drops (9.9\% for CIFAR-100 and 10.7\% for MiniImageNet). In the case of GSCv2, \sysname\ reveals a slight accuracy decline of 5.6\% compared to Oracle, while ANML+AIM shows a minor 0.2\% drop in accuracy relative to Oracle.

Although \sysname\ shows a slightly lower accuracy for GSCv2 than ANML+AIM, it still outperforms ANML+AIM by 4.1\% on average over all datasets. 
In addition, \sysname\ is essentially designed for edge devices to require drastically lower system resources (memory, latency, and energy) than the previous SOTA.
As explained in the following, the excessive resource overhead of ANML+AIM makes it unsuitable to operate on resource-constrained devices.

\begin{table}[t]
  \centering
  \caption{
  The required memory footprint and the compression ratio for the baselines and our system to perform CL during the meta-testing phase on the three datasets.
  }
  \label{tab:memory}
  \resizebox{1.015\columnwidth}{!}{%
  \begin{tabular}{ l  l   c c c c c c }
    \toprule 
     \textbf{Dataset} & \textbf{Metrics} & \textbf{Pretrained} & \textbf{ANML} & \textbf{OML+AIM} & \textbf{ANML+AIM} & \textbf{Oracle} & \textbf{\sysname}\\
        \cmidrule(l){0-7}
    \multirow{2}{*}{CIFAR-100} & Memory    & 39.69MB & 39.69MB & 834.1MB & 1,093MB & 39.93MB & \textbf{15.45MB} \\
     & Ratio & 27.5$\times$ & 27.5$\times$ & 1.3$\times$ & 1.0$\times$ & 27.4$\times$ & \textbf{70.8$\times$} \\
     \cmidrule(l){0-7}
    Mini-& Memory    & 474.5MB & 474.5MB & 1,051MB & 1,562MB & 475.0MB & \textbf{136.7MB} \\
     ImageNet & Ratio & 3.3$\times$ & 3.3$\times$ & 1.5$\times$ & 1.0$\times$ & 3.3$\times$ & \textbf{11.4$\times$} \\
     \cmidrule(l){0-7}
    \multirow{2}{*}{GSCv2} & Memory    & 10.16MB & 10.16MB & 135.2MB & 608.2MB & 10.20MB & \textbf{3.40MB} \\
     & Ratio & 59.9$\times$ & 59.9$\times$ & 4.5$\times$ & 1.0$\times$ & 59.6$\times$ & \textbf{178.7$\times$} \\
    \bottomrule
  \end{tabular}
  }
  % \vspace{-0.4cm}
\end{table}

\textbf{Peak Memory Footprint.}
We investigate the peak memory footprint required to perform CL.
Precisely, we measure the memory space required to perform backpropagation and to store rehearsal samples. The memory requirement to perform backpropagation consists of three components: (1) model memory that stores model parameters, (2) optimizer memory that stores gradients and momentum vectors, and (3) activation memory that is comprised of the intermediate activations (stored for reuse during backpropagation). Then, the memory requirement for rehearsal samples is included.

Table~\ref{tab:memory} shows the peak memory footprint for various baselines and our system. First, the AIM variants (OML+AIM and ANML+AIM) require an enormous memory footprint of 135.2-1,051 MB and 608.2-1,562 MB, respectively, as their AIM module has many parameters. This required memory easily exceeds the RAM size of embedded devices such as Pi 3B+ (i.e., 1 GB) and barely fits on Jetson Nano. Conversely, baseline systems such as Pretrained, ANML, and Oracle show modest memory requirements, which are around 10.16-10.20 MB for GSCv2, 39.7-39.9 MB for CIFAR-100, and 474.5-475.0 MB for MiniImageNet. However, as shown earlier, Pretrained and ANML methods are not highly accurate, and Oracle does not support CL. In contrast, \sysname\ shows the impressive results that it only requires 15.45 MB for CIFAR-100, 136.7 MB for MiniImageNet, and 3.40 MB for GSCv2, demonstrating a very high compression rate of 70.8$\times$, 11.4$\times$, and 178.7$\times$ compared to ANML+AIM, respectively. 
Compared to Oracle, \sysname\ shows a tight range of the compression (2.5-3.5$\times$), indicating that we can estimate the compression gain within this range agnostic to the dataset.

\begin{figure}[t]
  \centering
  \subfloat[Latency (CIFAR-100)]{
    \includegraphics[width=0.234\textwidth]{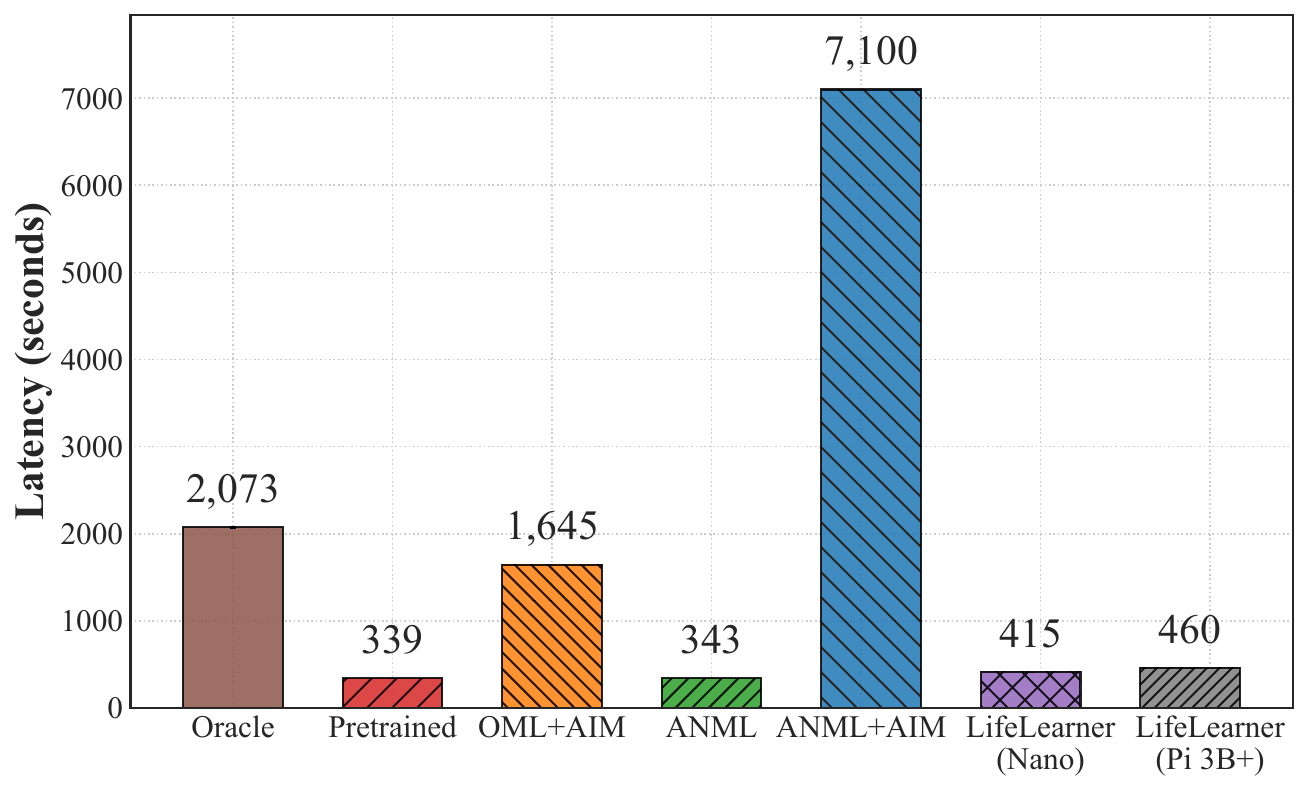}
  \label{subfig:cifar_latency}}
  \subfloat[Energy (CIFAR-100)]{
    \includegraphics[width=0.234\textwidth]{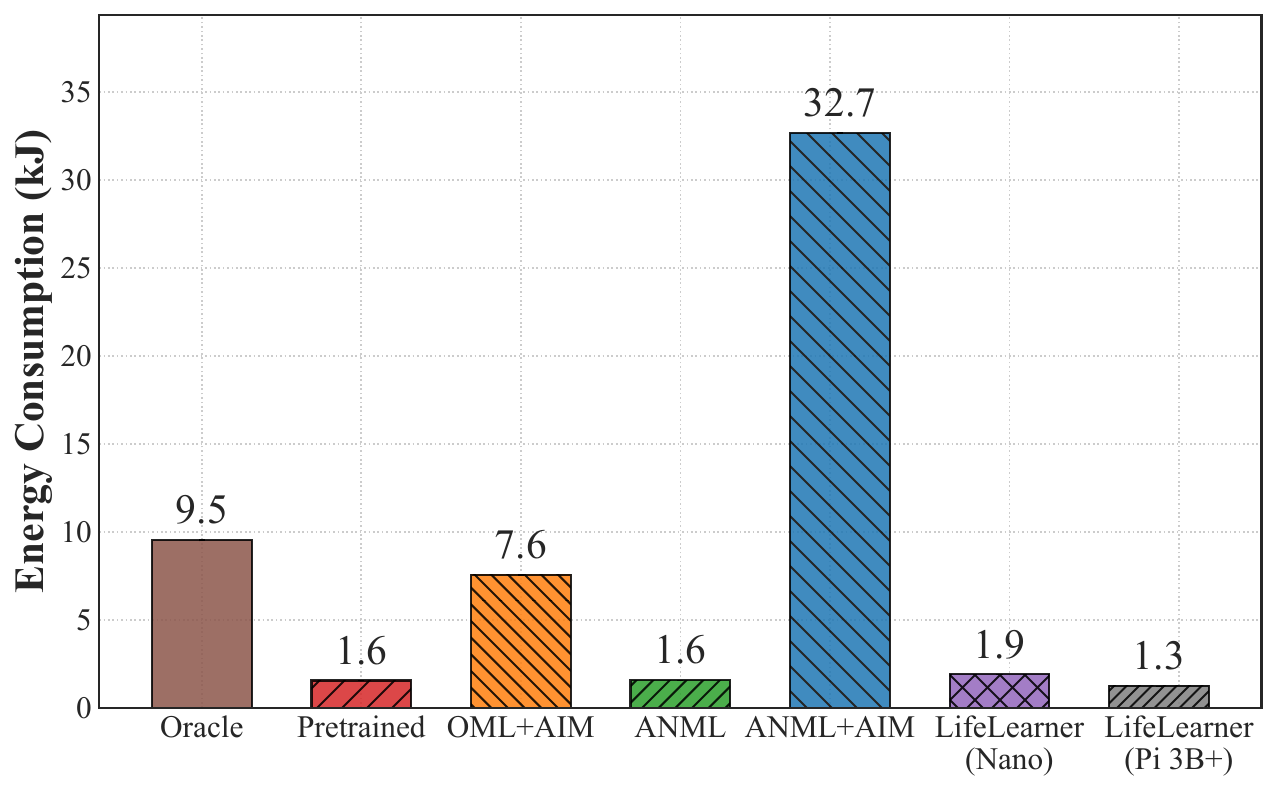}
  \label{subfig:cifar_energy}}
  \hrulefill
  \centering
  \subfloat[Latency (MiniImageNet)]{
    \includegraphics[width=0.234\textwidth]{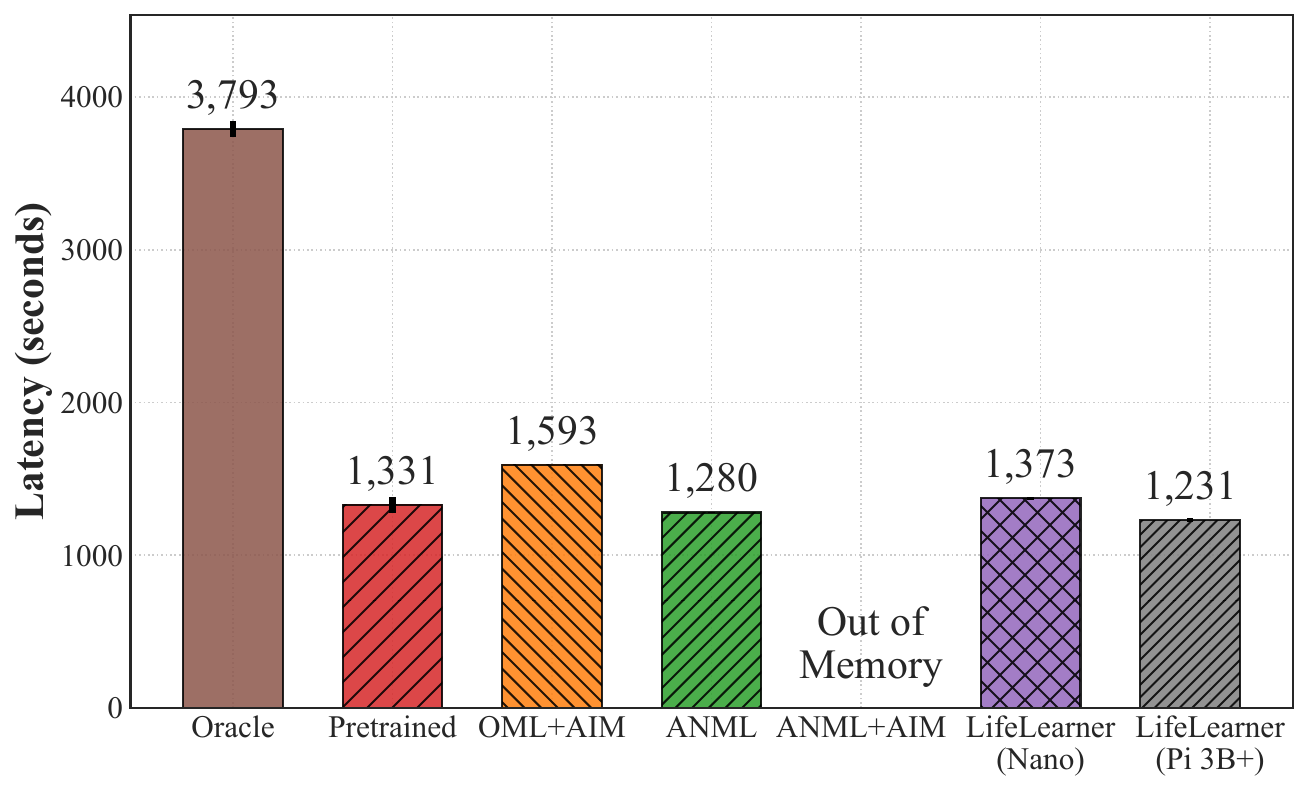}
  \label{subfig:miniimagenet_latency}}
  \subfloat[Energy (MiniImageNet)]{
    \includegraphics[width=0.234\textwidth]{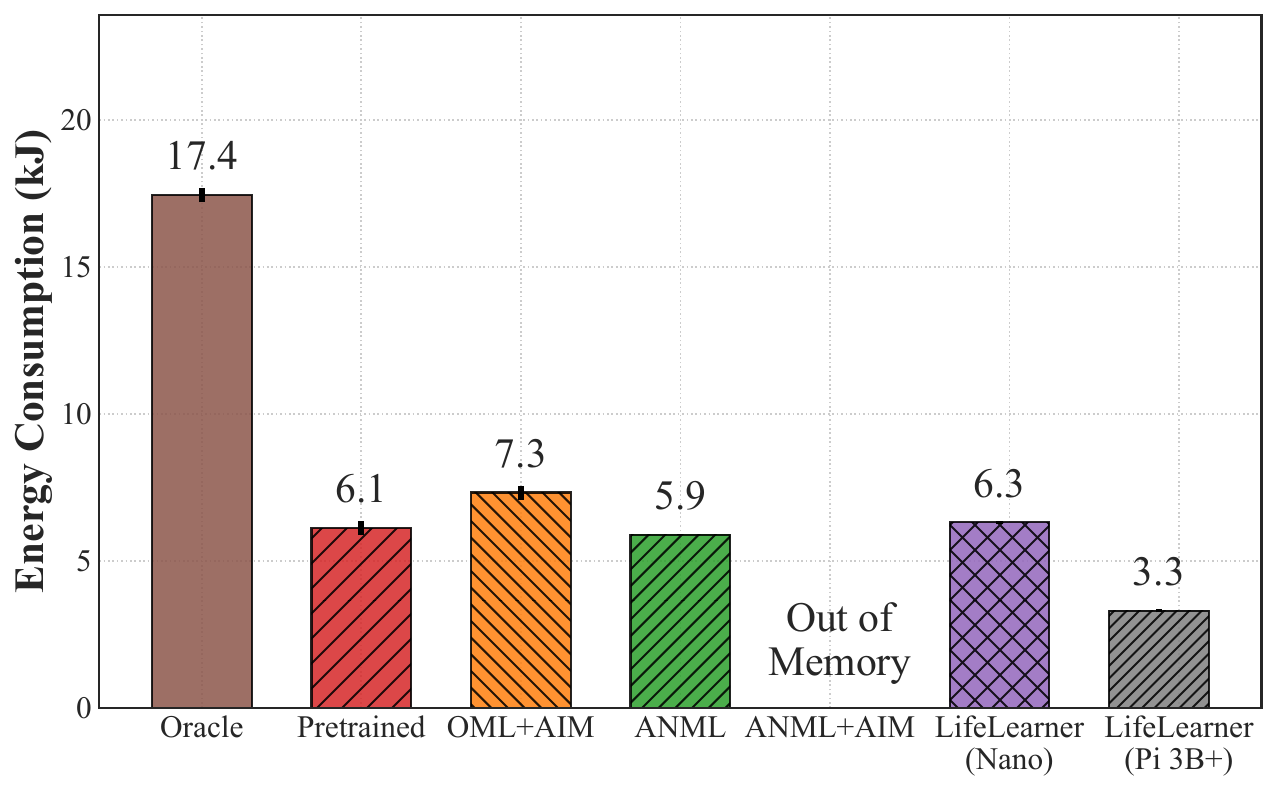}
  \label{subfig:miniimagenet_energy}}
  \hrulefill
  \centering
  \subfloat[Latency (GSCv2)]{
    \includegraphics[width=0.234\textwidth]{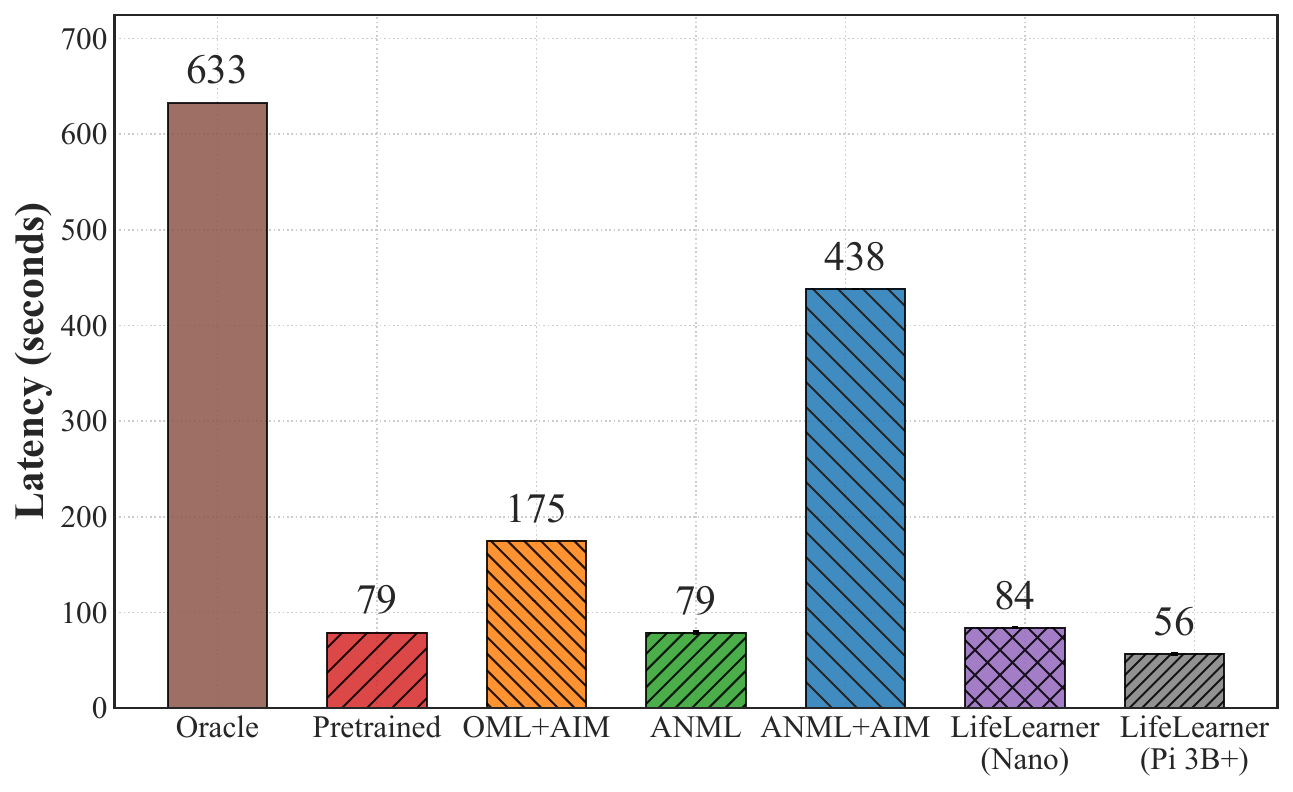}
  \label{subfig:gscv2_latency}}
  \subfloat[Energy (GSCv2)]{
    \includegraphics[width=0.234\textwidth]{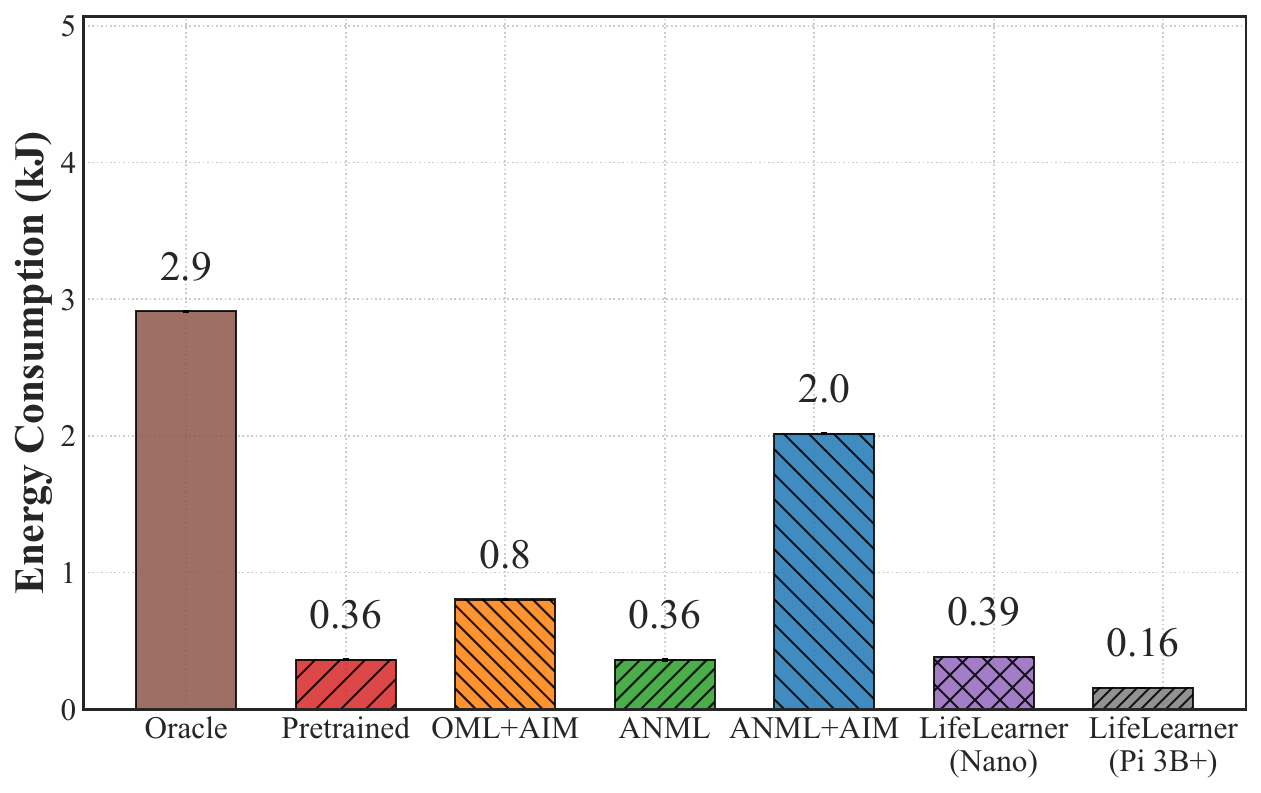}
  \label{subfig:gscv2_energy}}
  \hrulefill
  \centering
  \subfloat{
    \includegraphics[width=0.468\textwidth]{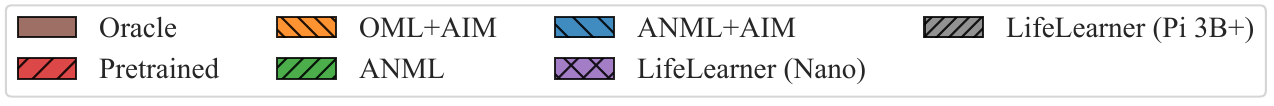}
  \label{subfig:labels}
  }
  \vspace{-0.2cm}
  \caption{
   The end-to-end latency and energy consumption of the baselines and \sysname\ to perform CL over all the given classes. All results are averaged over three runs with standard deviations.
  }
  \label{fig:latency and energy}
  % \vspace{-0.6cm}
\end{figure}

\textbf{End-to-end Latency \& Energy Consumption.}
We now examine the run-time system efficiency, i.e., end-to-end latency and energy consumption for the entire CL process, of our system and the baselines when deployed on the two embedded devices - Jetson Nano and Pi 3B+ as shown in Figure~\ref{fig:latency and energy}.
To obtain the end-to-end latency, we include: (1) the time to load a pretrained model, (2) the time to train the model continually over all the given classes one by one, and (3) the time to compress and decompress the latent representations using our compression method (i.e., sparse bitmap compression and PQ).

We first measure the end-to-end latency of our system and the baselines on Jetson Nano CPU to perform CL over all the given classes with 30 samples per class. 
As shown in Figures~\ref{subfig:cifar_latency},~\ref{subfig:miniimagenet_latency}, and~\ref{subfig:gscv2_latency}, \sysname\ enables a fast end-to-end latency (415 seconds for CIFAR-100, 1,373 seconds for MiniImageNet, and 84 seconds for GSCv2), which is 80.8-94.2\% reduction of latency compared to ANML+AIM (e.g., 7,100 seconds for CIFAR-100 and 438 seconds for GSCv2). Note that ANML+AIM often crashes from running out of memory on Jetson Nano due to its excessive memory requirements (as shown in Figures~\ref{subfig:miniimagenet_latency} and ~\ref{subfig:miniimagenet_energy}). Furthermore, compared to ANML which shares the same network architecture, \sysname\ introduces negligible overheads in terms of the overall latency (343s vs. 415s for CIFAR-100, 1,280s vs. 1,373s for MiniImageNet, and 79s vs. 84s for GSCv2). It is because although there exist some overheads on \sysname\ to perform the compression techniques like the sparse bitmap compression and PQ, the speed gains derived from using quantized neural weights and activations offset the overheads of compression techniques (refer to Section~\ref{subsec:ablation} for details). 
After having demonstrated the efficiency of \sysname\ on the Jetson Nano, we deployed our system on an even more resource-constrained device, Pi 3B+ (600-700 MB available memory). The end-to-end latency on Pi 3B+ largely stays similar to that on Jetson Nano as shown in Figure~\ref{fig:latency and energy}.

To measure the energy consumption, we first use Tegrastats on Jetson Nano to measure the power consumption. Then, we calculate the energy consumption by multiplying power consumption and the elapsed time for each end-to-end CL trial.
Similar to the latency results, Figures~\ref{subfig:cifar_energy},~\ref{subfig:miniimagenet_energy}, and~\ref{subfig:gscv2_energy} show that \sysname\ remarkably reduces the energy consumption by 80.9-94.2\% (1.9kJ vs. 32.7kJ for CIFAR-100 and 0.4kJ vs. 2.0kJ for GSCv2) compared to ANML+AIM. Moreover, compared to ANML, \sysname\ shows small overheads of the additional energy consumption (1.6kJ vs. 1.9kJ for CIFAR-100, 5.9kJ vs. 6.3kJ for MiniImageNet, and 0.36kJ vs. 0.39kJ for GSCv2).
In the case of Pi 3B+, it consistently consumes less energy than Jetson Nano. It is because while the end-to-end latency of the two embedded devices is similar, the power consumption profile on Pi 3B+ is lower than that on Jetson Nano, making Pi 3B+ a more energy-efficient option.
A YOTINO USB power meter is used to obtain the power consumption on Pi 3B+.

\textbf{Summary.}
\textit{Our result demonstrates that \sysname\ can effectively learn new classes in a continual manner based on only a few samples without experiencing catastrophic forgetting, i.e., it generalizes well to new samples of many classes unseen during the offline learning phase. Moreover, \sysname\ enables fast and energy-efficient CL on edge devices with significantly reduced memory footprint.}

\subsection{Ablation Study}\label{subsec:ablation}

We perform an ablation study to investigate the role of each component of our system by incrementally adding our proposed components on top of the baseline system (ANML): (1) rehearsal strategy with inner-and outer-loop optimization (Latent), (2) sparse bitmap compression (Latent+Bit), (3) PQ (Latent+PQ), and (4) quantization (\sysname).

\textbf{Effect of Rehearsal with Double-Loop Optimization.}
As shown in Table~\ref{tab:ablation}, we find that our proposed rehearsal strategy with double-loop optimization drastically improves the accuracy (compare ANML vs Latent). For example, Latent increases the accuracy of ANML by 10.6-28.4\% across all the datasets. Yet, Latent causes resource overheads on memory footprint, latency, and energy consumption compared to ANML, as Latent is a baseline CL system without our Compression Module.

\textbf{Effect of Compression and Hardware-aware Implementation.}
The results of various CL systems such as Latent+Bit, Latent+PQ, and Latent+Bit+PQ show that the proposed compression techniques for latent representations do not sacrifice the accuracy of the CL systems but reduce the overall memory footprint compared to Latent. Moreover, our Compression Module incurs small resource overheads in end-to-end latency and energy. 
Then, \sysname, which combines quantization of weights and activations accelerating the CL execution on hardware by exploiting efficient integer-based operations, shows excellent performance in all aspects: (1) outperforms ANML by a large margin (8.4-22.7\%) with a minor accuracy drop compared to Latent (0.9-5.7\%), (2) drastically reduces the memory footprint by 61.0-71.2\% compared to ANML and by 71.2-73.3\% compared to Latent, and (3) incurs minimal overheads of latency and energy over ANML (costs additional 56.6s and 0.3kJ on average, respectively) but still shows lower latency and energy than Latent (saves 47.9s and 0.2kJ on average, respectively).

\textit{Overall, the ablation study reveals that the co-utilization of the rehearsal strategy with double-loop optimization, Compression Module, and hardware-friendly implementation effectively makes \sysname\ more accurate and efficient.}

\begin{table}[t]
  \centering
  \caption{
  The comparison of \sysname\ and variants of rehearsal-based Meta CL methods for ablation study.
  }
  \label{tab:ablation}
  \resizebox{1.01\columnwidth}{!}{%
  \begin{tabular}{ l l   c c c c }
    \toprule 
     \textbf{Dataset} & \textbf{System} & \textbf{Accuracy} & \textbf{Memory} & \textbf{Latency} & \textbf{Energy} \\
        \cmidrule(l){0-5}
    \multirow{6}{*}{CIFAR-100}   & ANML           & 0.272 & 39.7 MB & 343.2s & 1.58kJ \\
                                 & Latent         & 0.452 & 53.9 MB & 432.5s & 1.99kJ \\
                                 & Latent+Bit     & 0.452 & 41.2 MB & 466.9s & 2.15kJ \\
                                 & Latent+PQ      & 0.448 & 41.8 MB & 437.1s & 2.01kJ \\
                                 & Latent+Bit+PQ  & 0.446 & 40.4 MB & 471.4s & 2.17kJ \\
                                  & \sysname       & \textbf{0.443} & \textbf{15.5 MB} & \textbf{414.7s} & \textbf{1.91kJ} \\
     \cmidrule(l){0-5}
                                 & ANML           & 0.327 & 474.5 MB & 1,280s & 5.89kJ \\
                                 & Latent         & 0.433 & 512.5 MB & 1,492s & 6.86kJ \\
                Mini-            & Latent+Bit     & 0.433 & 477.7 MB & 1,551s & 7.14kJ \\
                ImageNet         & Latent+PQ      & 0.430 & 483.0 MB & 1,501s & 6.90kJ \\
                                 & Latent+Bit+PQ  & 0.423 & 476.4 MB & 1,560s & 7.18kJ \\
                                 & \sysname       & \textbf{0.411} & \textbf{136.7 MB} & \textbf{1,373s} & \textbf{6.32kJ} \\
     \cmidrule(l){0-5}
    \multirow{6}{*}{GSCv2}       & ANML           & 0.429 & 10.2 MB & 78.6s & 0.36kJ \\
                                 & Latent         & 0.713 & 12.0 MB & 90.6s & 0.42kJ \\
                                 & Latent+Bit     & 0.713 & 10.4 MB & 90.8s & 0.42kJ \\
                                 & Latent+PQ      & 0.708 & 11.0 MB & 95.0s & 0.44kJ \\
                                 & Latent+Bit+PQ  & 0.707 & 10.3 MB & 95.2s & 0.44kJ \\
                                 & \sysname       & \textbf{0.656} & \textbf{3.40 MB} & \textbf{83.8s} & \textbf{0.39kJ} \\
    \bottomrule
  \end{tabular}
  }
  % \vspace{-0.3cm}
\end{table}

\begin{figure*}[t]
  \centering
  \subfloat[The Number of Samples per Class ($\mathcal{S}$)]{
    \includegraphics[width=0.31\textwidth]{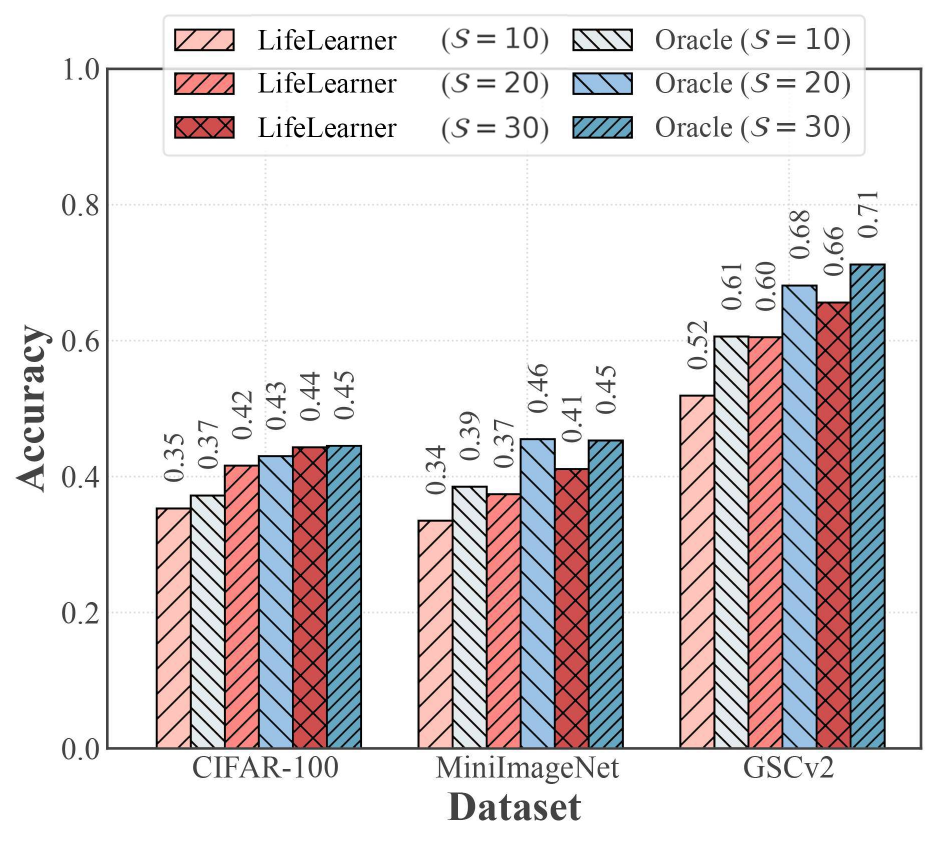}
  \label{subfig:param1}}
  \subfloat[The Number of Replay Epochs ($\mathcal{E}$)]{
    \includegraphics[width=0.31\textwidth]{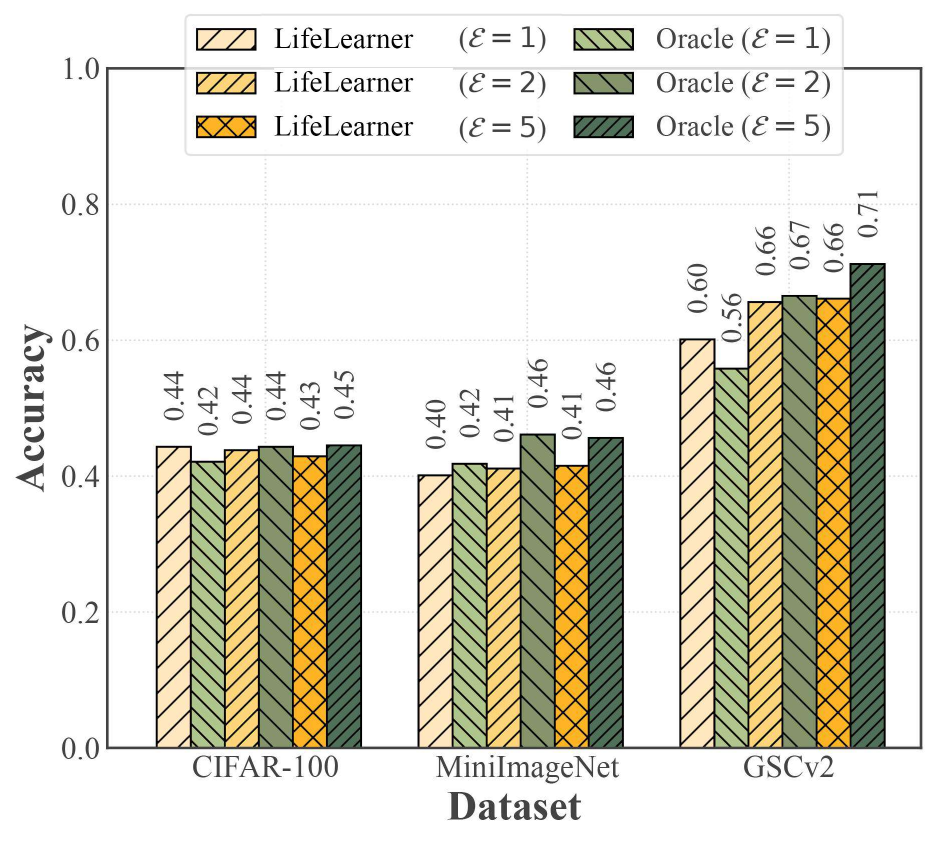}
  \label{subfig:param2}}
  \subfloat[The Sub-Vector Length ($\mathcal{L}$)]{
    \includegraphics[width=0.31\textwidth]{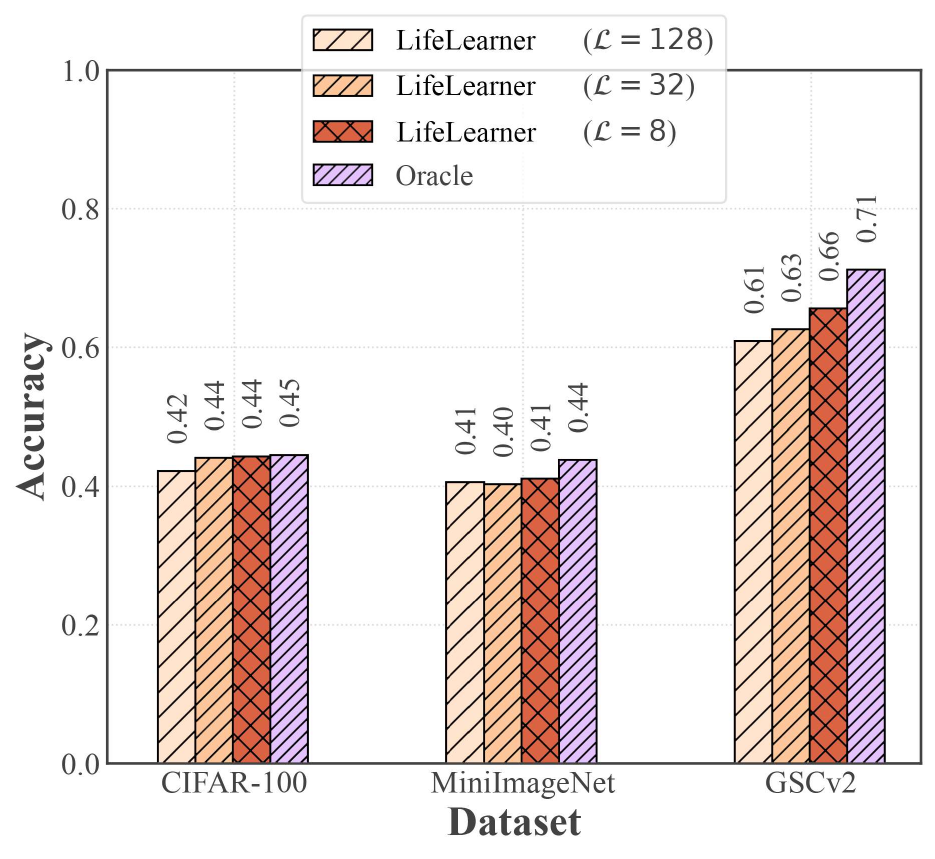}
  \label{subfig:param3}}
  \caption{
  The parameter analysis of \sysname\ for all the datasets according to the three parameters.
  }
  \label{fig:param analysis}
  % \vspace{-0.2cm}
\end{figure*}

\subsection{Parameter Analysis}\label{subsec:parameter analysis}

Next, we study the impact of the various hyper-parameters that could affect the performance of our system (see Figure~\ref{fig:param analysis}).

\textbf{The Number of Given Samples.} 
We first examine the accuracy of \sysname\ according to the number of given samples per class (ranging from 10 to 30) as it would directly affect labeling effort of users (see Figure~\ref{subfig:param1}). Apparently, the more samples are given for training, the higher the accuracy, which holds for both \sysname\ and Oracle. 
Even when only 10 samples per class are given to conduct training, the accuracy degradation of \sysname\ is relatively low (7-14\%), indicating that \sysname\ can still perform reasonably well under extreme data scarcity.
Also, the accuracy differences between \sysname\ and Oracle are small (e.g., 1-2\% for CIFAR-100, 1-3\% for MiniImageNet, and 5-9\% for GSCv2), demonstrating that \sysname\ achieves the similar accuracy of Oracle. With 30 given samples, the accuracy difference is minimal: 2.8\% on average (ranging from 1 to 5\%).

\textbf{The Number of Replay Epochs.}
We study to what extent the number of replay epochs affects the CL performance as more epochs incur larger latency and energy consumption.
Figure~\ref{subfig:param2} shows that the accuracy of \sysname\ converges after the first or the second replay epoch. However, Oracle requires at least two to five epochs to reach the convergence accuracy, which consumes much more training time and energy than our system (see Figure~\ref{fig:latency and energy}).
This result benefits us since replaying the rehearsal samples over one or two epochs is enough for \sysname\ to reach the converging accuracy, which helps decrease the system overheads.

\textbf{PQ Codebook's Sub-vector Length.}
We investigate the accuracy of \sysname\ according to the sub-vector length of the PQ codebook (the number of values per index) ranging from 8 to 128 as it affects the compression ratio of rehearsal samples. For CIFAR-100 and MiniImageNet, there is little difference according to the sub-vector length. In contrast, for GSCv2, we observe that the shorter the length of the sub-vector (i.e., lower compression rate), the higher the accuracy. These results inform us to select the largest sub-vector length that does not degrade accuracy.

\textit{These results show that with only 10-30 samples per class, \sysname\   achieve similar CL performance to Oracle, exhibit rapid convergence with small replay epochs (at most two), and accomplish a high compression rate for rehearsal samples.}

\subsection{MCU Deployment}\label{subsec:mcu deploy}
\textbf{TinyANML Architecture.} For the extremely resource constrained IoT devices like MCUs where on-chip memory of SRAM and Flash are typically a few hundred KB or 1 MB at most (an order of magnitude smaller than Jetson Nano and Pi 3B+ in terms of memory), the memory requirements of the MetaCL methods, including \sysname, are prohibitively large. Thus, we propose a small and accurate TinyANML architecture designed for MCUs with tiny memory by experimenting with various width modifiers~\cite{sandler_mobilenetv2_2018,lin_mcunet_2020,liberis_unas_2021}.
We identified widths of 0.2, 0.05, and 0.4 for the ANML architecture of CIFAR-100, MiniImageNet, and GSCv2, respectively.

\textbf{MCU Implementation and Results.}
Backbone represents an inference-only feature extractor based on TFLM. On top of that, our hardware-aware systems are added incrementally: (1) Backpropagation Engine (Tiny ANML) and (2) Compression Module (Tiny \sysname).
Table~\ref{tab:mcu_deploy} shows the MCU deployment results based on STM32H747 in terms of accuracy, SRAM, Flash, latency, and energy consumption to learn a class with ten samples when continually learning ten classes.

\textbf{Backpropagation Engine.}
As shown with Tiny ANML compared to inference-only Backbone, our Backpropagation Engine enables on-device CL with extremely small latency/energy overheads (e.g.,~579ms vs. 561ms and 134mJ vs. 128mJ for CIFAR-100) while requiring only an additional 100KB SRAM and 260KB Flash.

\textbf{Co-design of Our Algorithm and Hardware-aware System Implementation.} 
Tiny \sysname\ not only largely prevents accuracy degradation compared to its original \sysname\ (see Table~\ref{tab:ablation}) but also maintains higher accuracy than ANML despite Tiny \sysname's model size being 24.1-1839$\times$ smaller than ANML.
Tiny \sysname\ achieves significantly higher accuracy than Tiny ANML while having minimal resource requirements (e.g.,~181-281kB SRAM, 725-825kB Flash, 832-1,204ms latency, and 195-282mJ energy consumption), demonstrating the effectiveness of our proposed algorithm and hardware-aware system implementation on such an extremely resource-constrained device.

\textit{Note that it is infeasible to perform the ablation study to quantify the benefits of our design as in Section~\ref{subsec:ablation}. This is because other baselines with rehearsal strategy and prior works exhibit out-of-memory problems and only tiny \sysname\ could run on MCUs with severely limited memory.}

\begin{table}[t]
  \centering
  \caption{
  MCU deployment of the Backbone, tiny ANML, and tiny \sysname\ on STM32H747.}
  \label{tab:mcu_deploy}
  \resizebox{1.01\columnwidth}{!}{%
  \begin{tabular}{ p{1.6cm} p{2.4cm}   c c c c c }
    \toprule 
     \textbf{Dataset} & \textbf{System} & \textbf{Accuracy} & \textbf{SRAM} & \textbf{Flash} & \textbf{Latency} & \textbf{Energy} \\
        \cmidrule(l){0-6}
    \multirow{3}{*}{CIFAR-100}   & Backbone & - & 75kB & 428kB & 561ms & 128mJ \\
                                 & Tiny ANML & 0.176 & 185kB & 691kB & 579ms & 134mJ \\
                                 & Tiny \sysname & 0.393 & 236kB & 825kB & 832ms & 195mJ \\
     \cmidrule(l){0-6}
    \multirow{3}{*}{\parbox{1.5cm}{Mini-\\ImageNet}}& Backbone & - & 119kB & 329kB & 926ms & 221mJ \\
                                 & Tiny ANML & 0.112 & 224kB & 591kB & 944ms & 218mJ \\
                                 & Tiny \sysname & 0.301 & 281kB & 725kB & 1204ms & 282mJ \\
     \cmidrule(l){0-6}
    \multirow{3}{*}{GSCv2}       & Backbone & - & 81kB & 475kB & 956ms & 218mJ \\
                                 & Tiny ANML & 0.209 & 181kB & 738kB & 968ms & 223mJ \\
                                 & Tiny \sysname & 0.534 & 212kB & 806kB & 1160ms & 271mJ \\
    \bottomrule
  \end{tabular}
  }
  \vspace{-0.2cm}
\end{table}

\section{Discussion}\label{sec:discussion}

\textbf{Impact on Continual Learning.} 
We envision that \sysname\ could make CL a practical reality on embedded and IoT devices by leveraging meta-learning and rehearsal strategy with only a few samples. Such CL systems will allow DNNs to add new classes (e.g., adding new objects to an image recognition system, adding new keywords to a voice assistant) or new modalities (e.g., adding image recognition on top of a voice recognition authentication system) on the fly without relying on the cloud (i.e., no communication costs).
As one future direction, further optimizing \sysname\ to use stricter quantization such as 1, 2, or 4 bits will be interesting.

\textbf{Generalizability of \sysname.}
\sysname\ successfully works on three different datasets operating on two different modalities: image and audio, showing the generalizability of our framework. 
With the proliferation of smart spaces, such as smart homes and offices, \sysname\ can be used to learn the personal habits and preferences of users in order to control environmental conditions, such as temperature, humidity and lighting, with readings coming from thermometers, motion sensors and cameras on IoT devices. \sysname\ would enable this personalization and space adaptivity to happen in a data-efficient manner and to stay local to ensure privacy. Moreover, \sysname\ could be used on robot vacuum cleaners to enhance their adaptability, e.g.,~to continually learn to visually recognize new objects and thus avoid collisions.

The evaluation of other datasets and potentially other modalities, including various other sensor signals~\cite{pham_pros_mobicom22,cocoa_saeed} as mentioned above to further test the applicability of \sysname\ for learning continually for other real-world applications, is left as future work.

\textbf{Scalibility over Many Classes.}
The sample-wise compression ratio of \sysname\ is about 30$\times$, significantly reducing the memory overhead of adding many classes. It incurs only 1.68 MB, 6.16 MB, and 0.66 MB of memory when adding 100 classes with 30 samples per class for CIFAR-100, MiniImageNet, and GSCv2, respectively. Also, our scalar quantization and selective layer updates resolve scalability issues of latency as it incurs minimal latency overhead over ANML with fixed latency to learn new classes (see Tables~\ref{tab:ablation},~\ref{tab:mcu_deploy}).

\textbf{Feasibility of Labeling Samples.}
One of the key challenges of enabling realistic applications for CL is annotation difficulty by users. As conventional CL typically demands a few thousand labeled samples, it becomes almost infeasible for users to perform labeling (as discussed in Section~\ref{subsec:continual learning}). Instead, \sysname\ ameliorates this labeling burden by enabling data-efficient CL with 10-30 samples per class which are not impractical to label.

\textbf{Other Considerations.}
In this work, our evaluation demonstrated that \sysname\ achieves near-optimal CL performance, falling short by only 2.8\% accuracy compared to the upper bound system (Oracle). However, a higher accuracy (over 80-90\%) given fewer samples (less than 10-30 samples) would be desirable. Thus, it is worth investigating larger and more advanced model architectures specializing in the target problem and task, such as Transformers~\cite{attention_all_you_need,dosovitskiy2021vit}, to push the envelope of the upper bound testing accuracy of the challenging CL problem.

% \vspace{-0.2cm}
\section{Conclusions}\label{sec:conclusions}

We proposed \sysname, a hardware-aware meta CL system with adaptive fast-slow weights and resource-optimized compression for embedded and IoT platforms. 
\sysname\ outperforms all existing Meta CL methods by a large margin (approximating the upper bound method that performs training in i.i.d. setting) and demonstrates its potential applicability in real-world deployments.
Our efficient CL system opens the door to adaptive applications to run on embedded and IoT devices by allowing them to learn new tasks and adapt to the dynamics of the user and context.

\begin{acks}
This work is supported by a Google Faculty Award, ERC through Project 833296 (EAR), and Nokia Bell Labs through a donation
\end{acks}

\bibliographystyle{ACM-Reference-Format}
\bibliography{main}

\end{document}